\documentclass{article}
\usepackage{ijcai25}

\usepackage{times}
\usepackage{soul}
\usepackage{url}
\usepackage[hidelinks]{hyperref}
\usepackage[utf8]{inputenc}
\usepackage[small]{caption}
\usepackage{graphicx}
\usepackage{amsmath}
\usepackage{amsthm}
\usepackage{booktabs}
\usepackage{algorithm}
\usepackage{algorithmic}
\usepackage[switch]{lineno}

\DeclareMathOperator*{\argmax}{argmax}
\usepackage{subfig}
\usepackage{overpic}     
\usepackage{multirow} 
\usepackage{algorithm,algorithmic}
\usepackage{hyperref}
\usepackage{bm}
\usepackage{makecell}
\usepackage{flushend,cuted}
\usepackage{amssymb}
\usepackage{color}
\newtheorem{theorem}{Theorem}

\newtheorem{definition}{Definition}

\title{Negative Metric Learning for Graphs}

\author{
Yiyang Zhao$^1$
\and
Chengpei Wu$^1$\and
Lilin Zhang$^1$\And
Ning Yang$^1$\thanks{Corresponding author}\\
\affiliations
$^1$Sichuan University\\
\emails
\{zhaoyiyang0258, wuchengpei, zhanglilin\}@stu.scu.edu.cn,
yangning@scu.edu.cn
}

\begin{document}
\maketitle

\begin{abstract}
    Graph contrastive learning (GCL) often suffers from false negatives, which degrades the  performance on downstream tasks. The existing methods addressing the false negative issue usually rely on human prior knowledge, still leading GCL to suboptimal results. In this paper, we propose a novel Negative Metric Learning (NML) enhanced GCL (NML-GCL). NML-GCL employs a learnable Negative Metric Network (NMN) to build a negative metric space, in which false negatives can be distinguished better from true negatives based on their distance to anchor node. To overcome the lack of explicit supervision signals for NML, we propose a joint training scheme with bi-level optimization objective, which implicitly utilizes the self-supervision signals to iteratively optimize the encoder and the negative metric network. The solid theoretical analysis and the extensive experiments conducted on widely used benchmarks verify the superiority of the proposed method.
\end{abstract}

\section{INTRODUCTION}\label{introduction}

Graph contrastive learning (GCL) has emerged as a solution to the problem of data scarcity in the graph domain \cite{liu2022graph}. 
GCL employs an \textit{augmentation-encoding-contrasting} pipeline \cite{you2020graph} to obtain node- or graph-level representations without requiring labeled data.
The goal of GCL is to find a well-trained encoder (e.g., a two-layer GCN\cite{kipf2017semisupervised}) capable of generating informative representations capturing the underlying structure and features of an input graph, which can then be applied to downstream tasks. 

\begin{figure}[!t]
    \centering
    \subfloat[GCL with false negatives]{\includegraphics[scale=0.3]{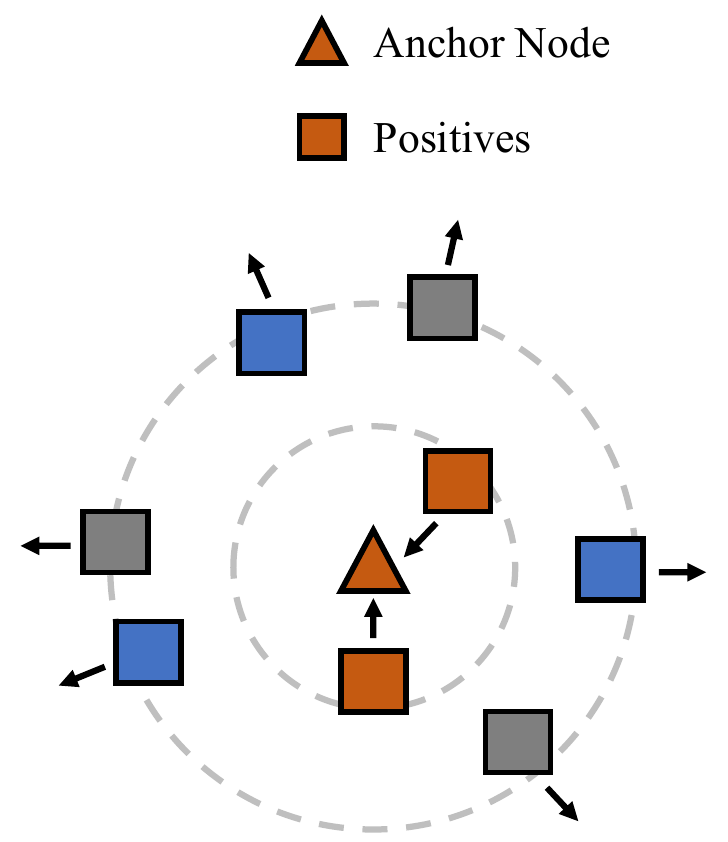}\label{fig:motivation-fn}}\hfil
    \subfloat[NML enhanced GCL]{\includegraphics[scale=0.3]{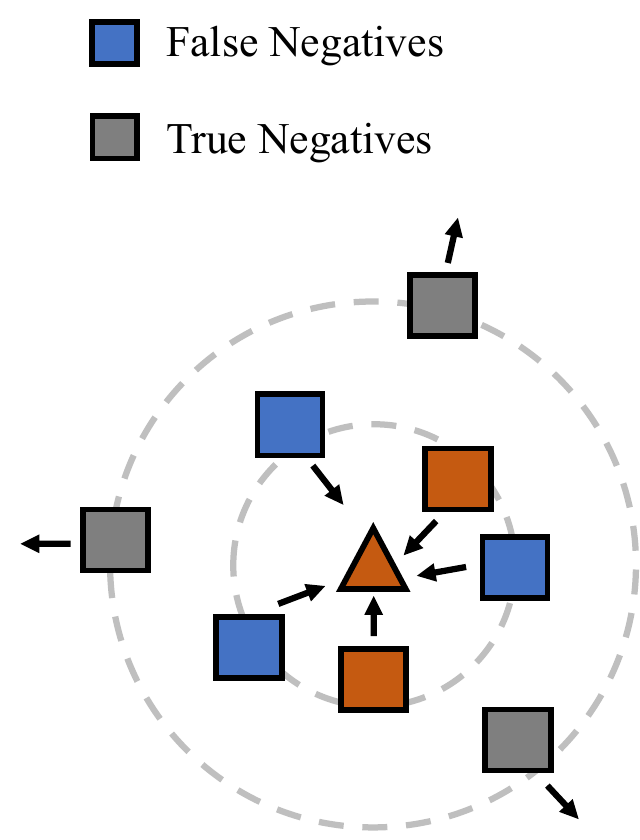}\label{fig:motivation-nml}}
    \caption{Illustration of GCL with false negatives and NML enhanced GCL.}
    \label{fig:motivation}
\end{figure}

The prevailing approach in GCL aims to maximize mutual information (MI) between different views, bringing positive sample embeddings closer together while pushing negative sample embeddings further apart \cite{velivckovic2018deep,Zhu2020deep}. For this purpose, InfoNCE, a lower bound of MI, is widely applied as the contrastive loss \cite{poole2019variational}. InfoNCE-based approaches often treat different views of the same node as positives, while those of different nodes as negatives. Such arbitrary and overly simplistic strategy makes the existing methods suffer from \textbf{false negatives}, i.e., the positive samples incorrectly treated as negatives. For example, in Fig. \ref{fig:motivation-fn}, the GCL with false negatives will result in that the embeddings of false negatives (blue squares) and true negatives (grey squares) are pushed away together from the anchor node (red triangle), which undermines the discriminative power of the learned embeddings, thereby reducing the performance on downstream tasks \cite{chuang2020debiased}. 

The existing methods usually address the issue of false negatives by weighting negative samples, which fall into two categories, hard-weight based 
\cite{zhang2022local,wang2024select,hu2021graph,liu2023b2,wuconditional,huynh2022boosting,fan2023neighborhood,han2023topology,sun2023progressive,yang2022region,li2023homogcl,liu2024seeking} and soft-weight based \cite{xia2022progcl,lin2022prototypical,hao2024towards,liu2024seeking,niu2024affinity,zhuo2024improving,wan2023boosting,chi2024enhancing,zhuo2024graph}. In particular, hard-weight based methods often assign binary weights to negative samples based on predefined criteria such as similarity threshold \cite{wuconditional,huynh2022boosting} or neighborhood distance \cite{li2023homogcl,liu2024seeking,zhang2022local}. For instance, in \cite{zhang2022local}, the first-order neighbors of an anchor node are considered false negatives. In contrast, soft-weight based methods relax the weight to $[0,1]$, of which one typical approach is first cluster the nodes, then determine the weight of negative samples based on the distance between the anchor node and the cluster where the negative sample resides \cite{lin2022prototypical}. However, both methods rely on simple prior knowledge to identify false negatives, which cannot guarantee precision or recall, still leading GCL to suboptimal results.

To overcome the above challenges, in this paper, we propose a novel \textbf{Negative Metric Learning (NML)} enhanced GCL (NML-GCL). The main idea of NML-GCL is to extend an InfoNCE-based GCL with a \textbf{Negative Metric Network (NMN)} to \textit{build a negative metric space where the less likely two nodes are true negatives of each other (and equivalently, the more likely two nodes are false negatives of each other), the closer their distance}. As shown in Fig. \ref{fig:motivation-nml}, compared with the true negatives, the false negatives' embeddings will be pulled closer to the anchor node under the guidance of the negative metric network. Essentially, this distance in the negative metric space can be considered a soft label indicating the node's status as a negative or positive sample with respect to the anchor. However, in the situation of self-supervised learning, the training of the negative metric network is difficult because there is a lack of explicit supervision signals regarding on negative/positive samples. To address this issue, inspired by the idea of self-training \cite{wei2020theoretical}, we propose a \textit{joint training scheme that can iteratively update the graph encoder (GCN) and the negative metric network with a bi-level optimization objective}. During the bi-level optimization, the negative metric network is responsible for assigning soft labels to samples based on the embeddings output by the encoder, while the encoder adjusts itself in the next iteration based on these soft labels. It is noteworthy that the insight here is the self-supervision signals (i.e., the different views of an anchor node) not only explicitly supervise the training of the encoder, but also implicitly supervise the training of the negative metric network, which makes them able to help each other.

We further provide a solid theoretical analysis of our proposed NML-GCL, revealing the connection between the negative metric network and the graph encoder. Specifically, we prove that: (1) our NML can enhance the GCL with a tighter lower bound of mutual information (MI) compared to traditional InfoNCE loss, and (2) by maximizing the tighter lower bound of MI, the joint training of the encoder and the negative metric network can mutually reinforce each other, leading to simultaneous improvements. Our contributions are summarized as follows:
\begin{itemize}
	\item[(1)] We propose a novel GCL framework NML-GCL. NML-GCL employs a learnable negative metric network to build a negative metric space, in which false negatives can be distinguished better from true negatives based on their distance to anchor node.
	
	\item[(2)] To overcome the lack of explicit supervision signals for NML, we propose a joint training scheme with bi-level optimization objective, which implicitly utilizes the self-supervision signals to iteratively optimize the encoder and the negative metric network.
	
	\item[(3)] Furthermore, we provide a solid theoretical justification of NML-GCL, by proving that due to NML, NML-GCL can approximate the MI between contrastive views with a tighter lower bound than traditional InfoNCE loss, which leads to the superiority of NML-GCL to the existing GCL methods.
 
	\item[(4)] The extensive experiments conducted on real-world datasets demonstrate the superiority of NML-GCL, in terms of the performance of the downstream tasks and the identifying of false negatives. 
\end{itemize}

Due to space limitations, related work is presented in Appendix \ref{related_work}.

\section{PRELIMINARIES}\label{preliminaries}
A graph is denoted by $\mathcal{G}=(\mathcal{V}, \mathcal{E})$, where $\mathcal{V}=\{i\}_{i=1}^N$, $\mathcal{E} \subseteq  \mathcal{V} \times \mathcal{V}$ represent the set of $N$ nodes and the set of edges respectively. 
Let $\mathbf{A} \in \{0,1\}^{N \times N}$ denote the adjacency matrix and $\mathbf{X} \in \mathbb{R}^{N \times F}$ be the node attribute matrix, where cell $a_{ij}$ at $i$-th row and $j$-th column of $\mathbf{A}$ is $1$ if $(i,j) \in \mathcal{E}$, otherwise 0, $\mathbf{x}_i \in \mathbb{R}^F$ is the $i$-th row of $\mathbf{X}$ representing the attribute vector of node $i$, and $F$ is the dimensionality.

\subsection{Graph Contrastive Learning}
GCL follows an \textit{augmentation-encoding-contrasting} mechanism basically. In the augmentation stage, GCL creates contrastive views preserving invariant structural information and feature information, by perturbing original graphs, e.g., edge masking \cite{Rong2020DropEdge,you2020graph}, node feature perturbation \cite{you2020graph}, or graph diffusion \cite{hassani2020contrastive}. In the encoding stage, GCL employs a GCN as encoder to generate the node embeddings, which is usually defined as 
\begin{equation}
    \mathbf{H}^{(k)} = \sigma(\tilde{\mathbf{A}}\mathbf{H}^{(k-1)}\mathbf{W}^{(k)}),
    \label{Eq:GCN}
\end{equation}
where $\mathbf{H}^{(k)}$ is the node embedding matrix at layer $k$, $\tilde{\mathbf{A}}$ is the normalized adjacency matrix, and $\mathbf{W}^{k}$ is the trainable weight matrix of layer $k$.

In the contrasting stage, the encoder is optimized by maximizing the agreement between the contrastive views, which is usually implemented with the InfoNCE loss \cite{poole2019variational} defined as:
\begin{equation}\label{eq:infonce}
\mathcal{L}_{\text{InfoNCE}}=\mathbb{E}_{i \in \mathcal{V}} \Big[ - \log\frac{e^{\theta(\mathbf{u}_{i},\mathbf{v}_{i})/\tau}}{e^{\theta(\mathbf{u}_{i},\mathbf{v}_{i})/\tau} + \sum^N_{j=1, j\neq i} e^{\theta(\mathbf{u}_{i},\mathbf{v}_{j})/\tau}} \Big ],
\end{equation}
where $\mathbf{u}_{i}$ and $\mathbf{v}_{i}$ are the embeddings of node $i$ in contrastive views $\mathcal{G}_{U}$ and $\mathcal{G}_{V}$, respectively, $\theta(\cdot,\cdot)$ is a similarity function (e.g., cosine), $\tau$ is a temperature parameter.

\subsection{False Negatives}
Let $\mathcal{S}_i$ be the negative sample set of an anchor node $i$ generated by a sampling strategy. Then the false negatives of anchor node $i$ with respect to $\mathcal{S}_i$ can be defined as
\begin{definition}[False Negatives] Node $j$ is a false negative of anchor node $i$ if $j \in \mathcal{S}_i$ and $Y(j) = Y(i)$, where $Y(\cdot)$ be the oracle label function unknown in advance.
\label{Def:false-negative}
\end{definition}
The idea of the above definition is that a false negative $j$ is a positive ($Y(j) = Y(i)$), but is treated as a negative ($j \in \mathcal{S}_i$), since the node labels are unobserved (i.e., $Y(\cdot)$ is unknown). In contrast, the true negatives are the nodes with labels different to the label of the anchor node. 

\begin{figure*}
    \centering
    \includegraphics[width=0.85\linewidth]{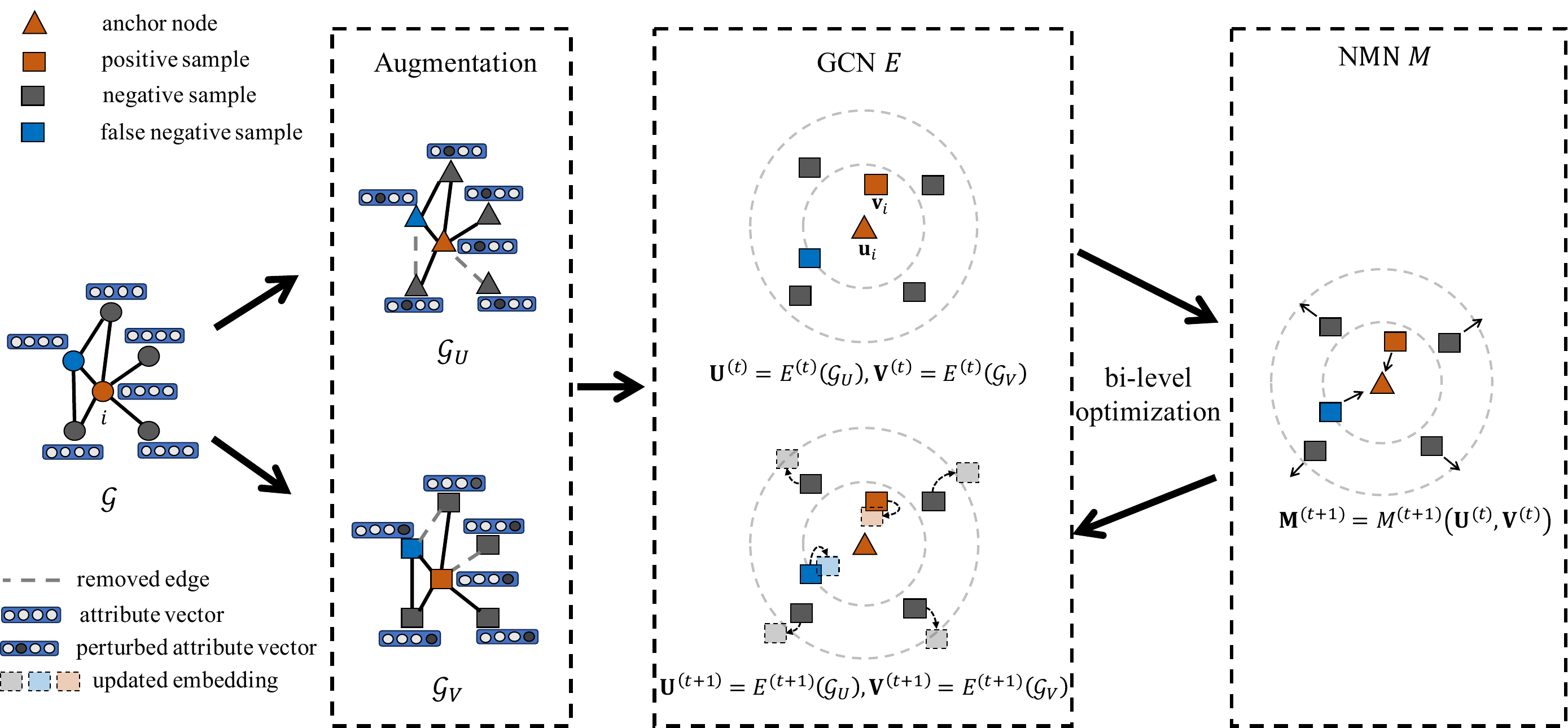}
    \caption{Overview of NML-GCL. First, NML-GCL generates two contrastive views $\mathcal{G}_U$ and $\mathcal{G}_V$ based on the initial graph $\mathcal{G}$. Then NML-GCL employs a bi-level optimization to iteratively training GCN $E$ and NMN $M$. In $(t+1)$-th iteration, NML-GCL first updates $M$ to $M^{(t+1)}$ under the guidance of GCN $E^{(t)}$ in previous iteration, then uses $M^{(t+1)}$ to obtain the new negative metric matrix $\mathbf{M}^{(t+1)}$ based on the node embeddings $\mathbf{U}^{(t)}$ and $\mathbf{V}^{(t)}$ in contrastive views, and finally, updates GCN $E$ to $E^{(t+1)}$ under the guidance of $\mathbf{M}^{(t+1)}$.}
    
%    First, two views $\mathcal{G}_U$ and $\mathcal{G}_V$ are generated from the initial graph $\mathcal{G}$. Next, the views are fed into a parameter-sharing GCN encoder $E$ at time $t$, i.e., $E^{(t)}$, to obtain the node embeddings $\mathbf{U}$ and $\mathbf{V}$. And then, positive and negative pairs are selected for contrasting, in which the negative ones are relabeled by the soft labels assigned by negative metric (NM) network $M$. At last, following a bi-level optimization, $M$ will be first updated to $M^{(t+1)}$ based on $\mathbf{U}$ and $\mathbf{V}$ produced by $E^{(t)}$, and then $E$ will be updated to $E^{(t+1)}$ through a contrastive loss relabeled by $M^{(t+1)}$. In the embedding space defined by $E^{(t+1)}$, the distance between false negative and anchor node can be relatively shortened compared with true negatives.}
    \label{fig:framework}
\end{figure*}

\section{METHODOLOGY}\label{method}
\subsection{Overview}
Fig. \ref{fig:framework} presents an overview of our proposed NML-GCL. From Fig. \ref{fig:framework} we can see that in the augmentation stage, NML-GCL first generates two contrastive views $\mathcal{G}_U$ and $\mathcal{G}_V$, each of which perturbs both the graph topology and the attributes of nodes of the original graph $\mathcal{G}$.
And then, the two augmentations $\mathcal{G}_U$ and $\mathcal{G}_V$ are sent to the GCN encoder $E$ defined in Equation (\ref{Eq:GCN}) to obtain the two node embedding matrices $\mathbf{U} = E(\mathcal{G}_U) \in \mathbb{R}^{N \times d}$ and $\mathbf{V} = E(\mathcal{G}_V) \in \mathbb{R}^{N \times d}$ for contrastive learning, where $d$ is the embedding dimensionality. 
In the contrastive learning stage, for an anchor node $i \in \mathcal{V}$, $(\mathbf{u}_i, \mathbf{v}_i)$, which are $i$-th rows of $\mathbf{U}$ and $\mathbf{V}$ representing the embeddings of $i$ in views $\mathcal{G}_U$ and $\mathcal{G}_V$ respectively, is selected as the positive pair, while $\{(\mathbf{u}_i, \mathbf{v}_j)\}_{j \in \mathcal{V}}$ as the negative pairs. Note that here in our NML-GCL, $(\mathbf{u}_i, \mathbf{v}_i)$ is considered as both positive and negative pair, so that their embedding distance can be adjusted by negative metric network (NMN) together with other negative pairs in a unified way. 

NML-GCL employs a bi-level optimization to jointly train GCN $E$ and negative metric network $M$ in an iterative fashion. In $(t+1)$-th iteration, NML-GCL first updates $M$ to $M^{(t+1)}$ with respective to the contrastive node embeddings $\mathbf{U}^{(t)} $ and $\mathbf{V}^{(t)} \in \mathbb{R}^{N \times d}$ generated by GCN $E^{(t)}$ in previous iteration. Then NML-GCL uses $M^{(t+1)}$ to obtain the new negative metric matrix $\mathbf{M}^{(t+1)} \in \mathbb{R}^{N \times N}$. At last, GCN $E$ is updated to $E^{(t+1)}$ with respective to $\mathbf{M}^{(t+1)}$. In the negative metric matrix $\mathbf{M}$, the cell at $i$-th row and $j$-th column $m_{ij}$ measures how likely node $j$ is a negative of node $i$. As mentioned before, $m_{ij}$ can be regarded as a soft label of $j$ indicating its status as a negative sample with respective to the anchor node $i$.

\subsection{Negative Metric Learning}\label{sec:NML}
To deal with the false negatives in GCL, we introduce Negative Metric Learning (NML) to learn a negative metric network $M$. The negative metric network $M$ captures the distance between two nodes from different views in the negative metric space, which reflects the probability they are negatives of each other. In particular, $M$ consists of an MLP and a normalizing layer, which is defined as
\begin{equation}
\begin{aligned} 
m^{'}_{ij} &= \text{MLP}(\mathbf{u}_i, \mathbf{v}_j), \\
m_{ij} &= \frac{e^{m^{'}_{ij}}}{ \sum_{k \in \mathcal{V}} e^{ m^{'}_{ik}}}, 
\end{aligned} 
\label{eq:m}
\end{equation}
where $m^{'}_{ij}$ is the distance between $\mathbf{u}_i$ and $\mathbf{v}_j$ in the negative sample space and $m_{ij} \in [0,1]$ is the normalized distance w.r.t. $\mathbf{u}_i$ satisfying $\sum_{j}m_{ij} = 1$. Therefore, $m_{ij}$ can be regarded as the probability (soft label) that $j$ is a negative of $i$.

We expect $M$ outputs smaller $m_{ij}$ for a false negative $j$ of an anchor node $i$. As mentioned before, however, there are no explicit supervision signals telling us false negatives. To overcome this issue, we introduce the similarity $\theta (\mathbf{u}_i,\mathbf{v}_i)$ induced by the GCN $E$ as the surrogate supervision signals for the training of $M$, by which the self-supervision signals (i.e., the supervision offered by the fact $(\mathbf{u}_i, \mathbf{v}_i)$ is positive pair) are transferred to the training of $M$. $\theta$ can be any qualified similarity measures, e.g., cosine. Based on this idea, the optimization objective of $M$ can be formulated as 
\begin{equation}
\min_{M} \text{ } \mathbb{E}_{i \in \mathcal{V}} \text{ } \mathcal{L}_{\text{NML}}^{(i)}, 
\end{equation}
where 
\begin{equation}\label{eq:cons_loss}
\mathcal{L}_{\text{NML}}^{(i)} = - \log \frac{ e^{\theta (\mathbf{u}_i,\mathbf{v}_i) / \tau} }{ e^{\theta (\mathbf{u}_i,\mathbf{v}_i) / \tau}  + (N-1) \sum_{j \in \mathcal{V}} m_{ij} e^{\theta (\mathbf{u}_i,\mathbf{v}_j) / \tau} }
\end{equation}
is a contrastive loss of anchor node $i$ with the temperature parameter $\tau$. It is obvious that when minimizing $\mathcal{L}_{\text{NML}}^{(i)}$ w.r.t. $M$, the bigger $\theta (\mathbf{u}_i,\mathbf{v}_j)$, the smaller $m_{ij}$. Note that, compared with traditional InfoNCE loss defined by Equation (\ref{eq:infonce}) where each $j \ne i$ is treated as negatives of $i$ with constant weight $1$, in Equation (\ref{eq:cons_loss}) all nodes $j \in \mathcal{V}$, even $i$ itself, are treated as potential negatives of $i$, and the weight of a negative becomes a learnable metric $m_{ij}$ induced by the negative metric network $M$.

To further clarify how the self-supervised signals implicitly supervise the training of $M$, we transform the loss defined by Equation (\ref{eq:cons_loss}) to an approximate hinge loss via the following derivation:
\begin{small}
\begin{equation} \label{Eq:margin}
\begin{aligned} 
\mathcal{L}_{\text{NML}}^{(i)}
=& \log \Big( 1 + (N-1) \sum_{j \in \mathcal{V}} m_{ij}  e^{\theta (\mathbf{u}_i,\mathbf{v}_j) - \theta (\mathbf{u}_i,\mathbf{v}_i)} \Big) \\
=& \log \Big( e^0 + \sum_{j \in \mathcal{V}} e^{\theta (\mathbf{u}_i,\mathbf{v}_j) - \theta (\mathbf{u}_i,\mathbf{v}_i) + \log \big( (N-1) m_{ij}\big) } \Big) \\
\approx & \max \big\{ 0, \big\{ \theta (\mathbf{u}_i,\mathbf{v}_j) - \theta (\mathbf{u}_i,\mathbf{v}_i) + \log \big( (N-1) m_{ij}\big) \big\}_{j \in \mathcal{V}} \big\} \\
=& \max \big\{ 0,  \log \big((N-1) m_{ii} \big), \\
& \{ \theta (\mathbf{u}_i,\mathbf{v}_j) - \theta (\mathbf{u}_i,\mathbf{v}_i) + \log \big((N-1) m_{ij}\big) \}_{j \in \mathcal{V}, j\ne i} \big\},
\end{aligned}
\end{equation}
\end{small}
where the third line holds because $\log (e^{x_1} + e^{x_2} + ... + e^{x_n})$ $ \approx$ $ \max$ $ \{x_1, x_2, ..., x_n\} $ \cite{zhang2024empowering}. 

As we will demonstrate later, under the guidance of the self-supervised signal that $(\mathbf{u}_i, \mathbf{v}_i)$ is positive pair, we can train a GCN $E$ capable of generating node embeddings that satisfy $\theta (\mathbf{u}_i,\mathbf{v}_i) \ge \theta (\mathbf{u}_i,\mathbf{v}_j)$ for $j \ne i$. Therefore, Equation (\ref{Eq:margin}) indicates that once $E$ is trained, the minimization of $\mathcal{L}_{\text{NML}}^{(i)}$ w.r.t. $M$ implicitly requires smaller $m_{ii}$ due to the bigger $\theta (\mathbf{u}_i,\mathbf{v}_i)$, which leads to a higher probability, via smaller $m_{ii}$, to the event that $\mathbf{v}_i$ is a false negative of $\mathbf{u}_i$. In other words, via bigger $\theta (\mathbf{u}_i,\mathbf{v}_i)$, the self-supervision signals can strengthen $M$'s ability to recognize false negatives by enforcing a smaller $m_{ii}$.

\subsection{Bi-level Optimization}\label{sec:opt}
We have seen that the training of $M$ depends on a reliable $\theta (\mathbf{u}_i,\mathbf{v}_j)$ induced by $E$, and however, the training of $E$ is guided by the negative metric matrix output by $M$. Our idea to break this dilemma is to iteratively optimize them with following bi-level optimization:
\begin{equation}\label{eq:objective}
\begin{aligned}
\min_{E} \min_{M} \text{ } \mathbb{E}_{i \in \mathcal{V}}  \big[ \mathcal{L}_{\text{NML}}^{(i)} + \alpha \mathcal{L}_{\text{reg}}^{(i)} \big], 
\end{aligned}
\end{equation}
where $\mathcal{L}_{\text{reg}}^{(i)} $ is the regularization loss and $\alpha$ controls the weight of the regularization. $\mathcal{L}_{\text{reg}}^{(i)} $ is defined as
\begin{equation}\label{eq:kl_loss}
\mathcal{L}_{\text{reg}}^{(i)}= (N-1)\text{KL}(P_0 || P_i ), 
\end{equation}
where $\text{KL}(\cdot||\cdot )$ is KL-divergence, $P_i$ is the distribution over $\{ m_{ij} \}_{j\in \mathcal{V}}$ given $i$, i.e., $P_i(j) = m_{ij}$, and $P_0$ is a uniformly distribution, i.e., $P_0(j) = 1/N$. $\mathcal{L}_{\text{reg}}^{(i)} $ constrains the feasible region of the $i$-th row $\mathbf{m}_i$ of the negative metric matrix $\mathbf{M}$ to prevent $\mathbf{m}_i$ from becoming a one-hot vector.

During the $(t+1)$-th iteration of the bi-level optimization defined in Equation (\ref{eq:objective}), the inner minimization will result in optimal $\mathbf{M}^{(t+1)}$, which is supervised by $\theta (\mathbf{U}^{(t)},\mathbf{V}^{(t)})$ induced by the node embeddings offered by $E^{(t)}$, as described in Section \ref{sec:NML}. Now we want to answer two questions for the outer minimization. 

(1) How does $\mathbf{M}^{(t+1)}$ supervise the training of $E^{(t+1)}$ through $\mathcal{L}_{\text{NML}}^{(i)} $? Again according to Equation (\ref{Eq:margin}), the minimization of $\mathcal{L}_{\text{NML}}^{(i)} $ w.r.t. $E$ requires to minimize $\max \big\{0, \theta (\mathbf{u}^{(t+1)}_i,\mathbf{v}^{(t+1)}_j) - \theta (\mathbf{u}^{(t+1)}_i,\mathbf{v}^{(t+1)}_i) + \log \big((N-1) m^{(t+1)}_{ij}\big) \big\}$. For this purpose, $E$ has to be adjusted to make sure that $\theta (\mathbf{u}^{(t+1)}_i,\mathbf{v}^{(t+1)}_j)$ is smaller than $\theta (\mathbf{u}^{(t+1)}_i,\mathbf{v}^{(t+1)}_i)$ by at least $\log \big((N-1) m^{(t+1)}_{ij}\big)$. Obviously, the bigger $m^{(t+1)}_{ij}$, the smaller $\theta (\mathbf{u}^{(t+1)}_i,\mathbf{v}^{(t+1)}_j)$. In other words, here $m^{(t+1)}_{ij}$ supervises the training of $E^{(t+1)}$ by telling it how far it should push $\mathbf{v}^{(t+1)}_j$ away from $\mathbf{v}^{(t+1)}_i$. 

(2) What is the relation between $E^{(t+1)}$ and $E^{(t)}$? As the training of $E^{(t+1)}$ is supervised by $\mathbf{M}^{(t+1)}$, to answer this question, we need to take a lose look at $\mathbf{M}^{(t+1)}$. Obviously, for an anchor node $i \in \mathcal{V}$, the optimal $m_{ik}^{(t+1)}$ ($k \in \mathcal{V}$) satisfies:
\begin{equation}
\begin{aligned}
&\nabla_{ m_{ik}} \Big( \mathcal{L}_{\text{NML}}^{(i)} + \alpha  \mathcal{L}_{\text{reg}}^{(i)} \Big) \\
= &\frac{ e^{\theta (\mathbf{u}^{(t)}_i,\mathbf{v}^{(t)}_k)} }{ e^{\theta (\mathbf{u}^{(t)}_i,\mathbf{v}^{(t)}_i)} + \sum_{j=1}^N m^{(t+1)}_{ij} e^{\theta (\mathbf{u}^{(t)}_i,\mathbf{v}^{(t)}_i)} } - \frac{\alpha}{m^{(t+1)}_{ik}} = 0
\end{aligned}
\end{equation}
and
\begin{equation}
\sum_{k \in \mathcal{V} } m_{ik}^{(t+1)} = 1,
\end{equation}
which together lead to the optimal 
\begin{equation}\label{eq:opt_d}
\begin{aligned}
m_{ik}^{(t+1)} = \sigma \bigg(\alpha \cdot \frac{ e^{\theta (\mathbf{u}^{(t)}_i,\mathbf{v}^{(t)}_i)} + \sum_{j=1}^N m^{(t+1)}_{ij} e^{\theta (\mathbf{u}^{(t)}_i,\mathbf{v}^{(t)}_i)} }{ e^{\theta (\mathbf{u}^{(t)}_i,\mathbf{v}^{(t)}_k)} } \bigg). 
\end{aligned}
\end{equation}

From Equation (\ref{eq:opt_d}), we can see that $ m_{ik}^{(t+1)} $ is inversely proportional to the similarity $ \theta (\mathbf{u}^{(t)}_i,\mathbf{v}^{(t)}_k) $. This means that if $E^{(t)}$ pushes $\mathbf{v}_k$ from $\mathbf{v}_i$ (i.e., smaller $ \theta (\mathbf{u}^{(t)}_i,\mathbf{v}^{(t)}_k) $), which leads to bigger $ m_{ik}^{(t+1)} $, then $E^{(t+1)}$ will push $\mathbf{v}_k$ even farther from $\mathbf{v}_i$ under the supervision of $ m_{ik}^{(t+1)} $ (see the answer of Question (1)). This result shows that the training of $E$ can be regarded as a form of self-training \cite{wei2020theoretical}, where the supervision signal is provided by $E$ via $M$.

Finally, the above analysis also shows that due to the iterative updating, the negative metric network $M$, which captures the distance in negative metric space, and the encoder $E$, which generates discriminative embeddings, would reinforce each other during the bi-level optimization. As we will theoretically justify in the later, the bi-level optimization accurately maximizes the mutual information between contrastive views, which leads to $E$ and $M$ capable of distinguishing false negatives from true negatives. The complete training process is presented in Algorithm \ref{algorithm}, and its time complexity analysis is shown in Appendix \ref{time}.

\begin{algorithm}[!h]
    \renewcommand{\algorithmicrequire}{\textbf{Input:}}
	\renewcommand{\algorithmicensure}{\textbf{Output:}}
	\caption{Training process of our proposed NML-GCL}
    \label{algorithm}
    \begin{algorithmic}[1]
        \REQUIRE  A graph $\mathcal{G}$, an encoder $E$, a negative metric network $M$, the number of training epochs $T_\text{E}$ for outer minimization, the number of iterations $T_\text{M}$ for inner minimization;
        \ENSURE The optimal encoder $E$;
        \STATE Initialize parameters of $E$ and $M$;
        \FOR {$i=1,2, \cdots, T_\text{E}$}
            \STATE Randomly generate contrastive views $\mathcal{G}_U$ and $\mathcal{G}_V$ from $\mathcal{G}$;
            \STATE Generate node embeddings $\mathbf{U}$ and $\mathbf{V}$ by $E$;
            \STATE // Training negative metric network
            \STATE Freeze parameters of $E$;
            \FOR{$j=1,2, \cdots, T_\text{M}$}
                \STATE Update $M$ according to Equation (\ref{eq:objective});
            \ENDFOR
            \STATE // Training Encoder
            \STATE Freeze parameters of $M$;
                \STATE Update $E$ according to Equation (\ref{eq:objective});
        \ENDFOR    
    \end{algorithmic}
\end{algorithm}

\section{THEORETICAL ANALYSIS}\label{theory}
In this section, we first prove that compared with traditional InfoNCE-based GCL, our NML-GCL maximizes a tighter lower bound of the mutual information (MI) between contrastive views. Then we demonstrate that the maximization of the MI endows the encoder $E$ and the negative metric network $M$ with the ability to distinguish false negatives from true negatives. 

%can maximize a tighter lower bound of the mutual information (MI) between the two views on embedding level, i.e, $I(U; V)$, where $U$ and $V$ are random variables representing node embeddings in contrastive views $\mathcal{G}_{U}$ and $\mathcal{G}_{V}$, respectively. Then we will demonstrate that the maximization of $I(U; V)$ would lead to negative metric network $M$'s better modeling of negative sample space. 

\subsection{Tighter Lower Bound of MI}
Let $U$ and $V$ be random variables representing node embeddings in contrastive views $\mathcal{G}_{U}$ and $\mathcal{G}_{V}$, respectively, and $I(U; V)$ be their MI. Let $I_{\text{NML}} (U; V)$ and $I_{\text{NCE}}(U; V)$ denote the MI estimated by $\mathcal{L}_{\text{NML}}^{(i)}$ defined in Equation (\ref{eq:cons_loss}) and traditional InfoNCE loss $\mathcal{L}_{\text{InfoNCE}}$ defined in Equation (\ref{eq:infonce}), respectively. The following theorem shows that $I_{\text{NML}} (U; V)$ is a tighter lower bound of $I(U; V)$ than $I_{\text{NCE}}(U; V)$.
\begin{theorem}
$I (U; V) \ge I_{\text{NML}} (U; V)  \ge I_{\text{NCE}}(U; V)$, where $I_{\text{NML}}$ $ (U; V) = - \min_{M} \mathbb{E}_{i \in \mathcal{V}} [ \mathcal{L}_{\text{NML}}^{(i)} ] + C $, $I_{\text{NCE}}(U; V)$ $ = \mathcal{L}_{\text{InfoNCE}} + C $ and $C = \log N$.
\label{th:tighter}
\end{theorem}
The detailed proof of Theorem \ref{th:tighter} can be seen in Appendix \ref{proof-th1}. Basically, Theorem \ref{th:tighter} offers the rationality of NML-GCL's bi-level optimization by which NML-GCL achieves better generalizability on downstream tasks than traditional InfoNCE based GCL methods. 

\subsection{MI Maximization Facilitates NML}
Theorem \ref{th:tighter} shows that the minimization of $\mathcal{L}_{\text{NML}}^{(i)}$ in each iteration of the bi-level optimization defined in Equation (\ref{eq:objective}) approximately maximizes $I (U; V)$ by maximizing its tighter lower bound $I_{\text{NML}} (U; V)$. Now we further demonstrate that the maximization of $I (U; V)$ facilitates NML, enabling the encoder and the negative metric network to ultimately acquire the ability to distinguish between false negatives and true negatives in the negative metric space learned by NML. At first, the following theorem shows that an optimal $E$ can simulate the oracle label function $Y$ by appropriately ranking the distance of samples to a given anchor node in the negative metric space.
\begin{theorem}
If $E^* = \max_{E} \mathbb{E}_{\mathcal{G}_U, \mathcal{G}_V} I(U; V)$, then 
\begin{equation*}
\mathbb{E}_{i \in \mathcal{V}, j \in \mathcal{S}_i^{+}} [ d(\mathbf{z}_i, \mathbf{z}_j) ]  \le \mathbb{E}_{i \in \mathcal{V}, j \in \mathcal{S}_i^{-}}[ d(\mathbf{z}_i, \mathbf{z}_j)],
\end{equation*}
where $d(\cdot, \cdot)$ is a distance metric, $\mathcal{S}_i^{-}$ is the true negative set of $i$, $\mathcal{S}_i^{+}$ is the false negative set of $i$, and $\mathbf{z}_i$ represents the embedding of node $i$ in original graph $\mathcal{G}$ generated by encoder $E^*$. 
\label{th:relation}
\end{theorem}

The proof of Theorem \ref{th:relation} can be found in Appendix \ref{proof:th2}. Theorem \ref{th:relation} tells us that after maximizing $I(U; V)$, in the embedding space defined by $E^*$, the expected distance from false negatives to an anchor node is shorter than the expected distance from true negatives to the anchor node. In other words, $E^*$ can simulate the behavior of the oracle label function $Y$, since $d(\mathbf{z}_i, \mathbf{z}_j)$ approaches to $0$ if $Y(j) = Y(i)$, which explains why our NML-GCL is able to obtain node embeddings that are more discriminative in terms of class distinction. 

Theorem \ref{th:relation} together with Theorem \ref{th:tighter} and Equation (\ref{eq:opt_d}) justifies that $E$ and $M$ can iteratively reinforce each other via the bi-level optimization. Specifically, in one iteration, the inner minimization results in better $M$ that offers better supervision for the training of $E$; the outer minimization results in better $E$ that can induce  more reliable $\{ \theta (\mathbf{u}_i,\mathbf{v}_j) \}$ as the supervision for the training of $M$ in next iteration. Such positive feedback loop ensures the effectiveness of our NML-GCL.

\section{EXPERIMENTS}\label{experiment}

In this section, we conduct experiments to answer the following research questions (RQs):
\begin{itemize}
\item \textbf{RQ1:} Does NML-GCL outperform existing GCL methods on downstream tasks?
\item \textbf{RQ2:} How Negative Metric Network $M$ contributes to the performance of NML-GCL?
% \item[\textbf{RQ2:}] Can NML-GCL effectively distinguish false negative samples from true negative ones?
% \item[\textbf{RQ3:}] Can the theoretical analysis be experimentally verified?
\item \textbf{RQ3:} How do the hyper-parameters affect the performance of NML-GCL?
\end{itemize}
We conduct additional experiments, including experimental verification of theoretical analysis and the identification of false negatives, as detailed in Appendix \ref{addExp}.

\subsection{Experiment Settings}
\paragraph{Datasets.} We conduct experiments on six publicly available and widely used benchmark datasets, including three citation networks Cora, CiteSeer, PubMed \cite{yang2016revisiting}, two Amazon co-purchase networks (Photo, Computers) \cite{shchur2018pitfalls}, and one Wikipedia-based network Wiki-CS \cite{mernyei2020wiki}. The statistics of datasets are summarized in Table \ref{tab:datasets} in Appendix \ref{hyper-appendix}.

% \begin{table}[t]
%  \centering
%  \caption{Characteristics of the baseline methods and NML-GCL. }
%  \label{Tb:baselines}
%  \setlength{\tabcolsep}{0.05cm}
%  \begin{tabular}{l|c|c|c}
%  \toprule
%  \textbf{Method}&\textbf{Hard Weight}&\textbf{Soft Weight}&\textbf{Metric Learning}\\
%  \midrule
%   {BGRL \cite{thakoor2021bootstrapped}}&{}&{}&{}\\ 
%  {GRACE \cite{Zhu2020deep}}&{}&{}&{}\\ 
%  {MVGRL \cite{hassani2020contrastive}}&{}&{}&{}\\ 
% % \midrule
%  {LOCAL-GCL \cite{zhang2022local}}&{$\checkmark$}&{}&{}\\ 
%  {PHASES \cite{sun2023progressive}}&{$\checkmark$}&{}&{}\\ 
%  {HomoGCL \cite{li2023homogcl}}&{$\checkmark$}&{}&{}\\ 
% % \midrule
%  {GRACE+ \cite{chi2024enhancing}}&{}&{$\checkmark$}&{}\\ 
%  {ProGCL \cite{xia2022progcl}} &{}&{$\checkmark$}&{}\\ 
%  {GRAPE \cite{hao2024towards}} &{}&{$\checkmark$}&{}\\ 
% % \midrule
%  {NML-GCL}&{}&{$\checkmark$}&{$\checkmark$}\\ 
%  \bottomrule
%  \end{tabular}
% \end{table}

\begin{table*}[t]
  \centering
  % \setlength{\tabcolsep}{0.1cm}
  % \small
  % \renewcommand{\arraystretch}{0.85}
  \scalebox{0.85}{
    \begin{tabular}{lcccccc}
    \toprule
    \textbf{Method} & \textbf{Cora} & \textbf{CiteSeer} & \textbf{PubMed} & \textbf{Photo} & \textbf{Computers} & \textbf{Wiki-CS} \\
    \midrule
    BGRL & 82.22$\pm$0.51 & 72.27$\pm$0.26 & 79.41$\pm$0.23 & \underline{93.01$\pm$0.27} & 88.25$\pm$0.30 & 77.81$\pm$0.42 \\
    GRACE & 81.52$\pm$0.24 & 70.12$\pm$0.17 & 78.32$\pm$0.45 & 91.88$\pm$0.18 & 87.15$\pm$0.21 & 77.95$\pm$0.23 \\
    MVGRL & 81.15$\pm$0.16 & 70.46$\pm$0.23 & 78.63$\pm$0.27 & 92.59$\pm$0.34 & 87.66$\pm$0.25 & 78.52$\pm$0.15 \\
    \midrule
    LOCAL-GCL & 83.86$\pm$0.21 & 71.78$\pm$0.48 & 80.95$\pm$0.41 & 92.83$\pm$0.28 & \underline{88.69$\pm$0.50} & 78.98$\pm$0.52 \\
    PHASES & 82.19$\pm$0.37 & 70.49$\pm$0.40 & 81.08$\pm$0.62 & 92.74$\pm$0.37 & 87.80$\pm$0.44 & 79.61$\pm$0.29 \\
    HomoGCL & 83.19$\pm$1.03 & 70.11$\pm$0.79 & 81.02$\pm$0.68 & 92.31$\pm$0.36 & 87.82$\pm$0.48 & 78.18$\pm$0.45 \\
    \midrule
    GRACE+ & 82.84$\pm$0.18 & 70.57$\pm$0.53 & \underline{81.13$\pm$0.31} & 92.73$\pm$0.35 & 88.56$\pm$0.31 & 79.33$\pm$0.27 \\
    ProGCL & 82.04$\pm$0.27 & 70.63$\pm$0.22 & 78.14$\pm$0.43 & 92.71$\pm$0.29 & 87.90$\pm$0.27 & 78.21$\pm$0.41 \\
    GRAPE & \underline{83.88$\pm$0.04} & \underline{72.34$\pm$0.22} & OOM & 92.76$\pm$0.30 & 87.97$\pm$0.33 & \underline{79.91$\pm$0.16} \\
    \midrule
    NML-GCL & \textbf{84.93$\pm$0.14} & \textbf{73.37$\pm$0.13} & \textbf{82.10$\pm$0.30} & \textbf{93.36$\pm$0.21} & \textbf{89.43$\pm$0.25} & \textbf{80.32$\pm$0.18} \\
    \textit{(p\mbox{-}value)} & \textit{(9.88e\mbox{-}15)} & \textit{(1.89e\mbox{-}10)} & \textit{(1.25e\mbox{-}6)} & \textit{(4.58e\mbox{-}3)} & \textit{(5.55e\mbox{-}4)} & \textit{(4.07e\mbox{-}5)} \\
    \midrule
    w/o $M$ & ${82.82\pm0.07}$ & ${70.31\pm0.17}$ & ${80.14\pm0.14}$ & ${92.12\pm0.16}$ & ${87.76\pm0.19}$ & ${77.91\pm0.24}$ \\
    repl. cosine sim. & ${83.87\pm0.22}$ & ${71.18\pm0.24}$ & ${81.31\pm0.24}$ & ${92.80\pm0.22}$ & ${88.35\pm0.21}$ & ${79.13\pm0.14}$ \\
    \bottomrule
    \end{tabular}%
 }
  \caption{Node classification accuracy (\%) with standard deviation. 'OOM' means out of memory on a 24GB GPU. The best result is in bold, and the second best is underlined.}
  \label{tab:nodeclass}%
\end{table*}%

\paragraph{Baselines.} To validate the effectiveness of our NML-GCL, we compare it with state-of-the-art GCL methods including: 
\begin{itemize}
\item three methods without considering false negatives, BGRL \cite{thakoor2021bootstrapped}, GRACE \cite{Zhu2020deep}, and MVGRL \cite{hassani2020contrastive}, 

\item three methods dealing with false negatives with hard weight, LOCAL-GCL \cite{zhang2022local}, PHASES \cite{sun2023progressive}, HomoGCL \cite{li2023homogcl}, 

\item three methods dealing with false negatives with soft weight, GRACE+ \cite{chi2024enhancing}, ProGCL \cite{xia2022progcl}, and GRAPE \cite{hao2024towards}, where ProGCL and GRAPE determine the weights based on clustering. 
\end{itemize}
% The characteristics of the baseline methods are summarized in Table \ref{Tb:baselines}.

\paragraph{Evaluation Protocols.} We follow a two-stage evaluation protocol widely used by existing works \cite{zhang2021canonical,Zhu2020deep}. For each method, in the first stage, we generate node embeddings, while in the second stage, we evaluate the node embeddings in terms of the performance of node classification and node clustering. Specifically, for node classification, the classifier is implemented as a logistic regression. The node embeddings generated by a baseline method or NML-GCL is split into a training set, a validation set, and a testing set. The training set and validation set are used for the training and hyper-parameter tuning of the classifier, while testing set for evaluation in terms of accuracy. For node clustering, we use $k$-means algorithm to partition the node embeddings where $k$ is set to the number of classes in a dataset, and evaluate the clustering results by two widely used metrics, Fowlkes-Mallows Index (FMI) \cite{campello2007fuzzy} and Adjusted Rand Index (ARI) \cite{steinley2004properties}. 

\paragraph{Configurations.} In NML-GCL, the encoder $E$ is implemented as a two-layer GCN with embedding dimensionality $d=512$, and the NMN is implemented as an MLP with two hidden layers each of which consists 512 neurons. We apply Adam optimizer for all the GCL methods and the classifier. 
Following the approach of \cite{Zhu2020deep}, we adopt DropEdge \cite{Rong2020DropEdge} and FeatureMasking \cite{you2020graph} to generate contrastive views. The detailed hyper-parameter settings are shown in Table \ref{tab:datasets} in Appendix \ref{hyper-appendix}.

\begin{table*}[t]
  \centering
  \scalebox{0.85}{
    \begin{tabular}{l|cc|cc|cc|cc|cc|cc}
    \toprule
    \multirow{2}{*}{\textbf{Method}} & \multicolumn{2}{c|}{\textbf{Cora}} & \multicolumn{2}{c|}{\textbf{CiteSeer}} & \multicolumn{2}{c|}{\textbf{PubMed}} & \multicolumn{2}{c|}{\textbf{Photo}} & \multicolumn{2}{c|}{\textbf{Computers}} & \multicolumn{2}{c}{\textbf{Wiki-CS}} \\
    \cmidrule{2-13}
    & \textbf{FMI} & \textbf{ARI} & \textbf{FMI} & \textbf{ARI} & \textbf{FMI} & \textbf{ARI} & \textbf{FMI} & \textbf{ARI} & \textbf{FMI} & \textbf{ARI} & \textbf{FMI} & \textbf{ARI} \\
    \midrule
    BGRL & 52.74 & 42.82 & 46.37 & 34.27 & 54.40 & 28.44 & \underline{62.59} & 50.34 & 45.37 & 31.68 & 36.16 & 27.96 \\
    GRACE & 53.24 & 43.68 & 45.36 & 33.71 & 51.47 & 24.63 & 57.27 & 47.42 & 46.83 & 35.22 & 37.61 & 28.24 \\
    MVGRL & 55.49 & 46.68 & 47.43 & 37.82 & 51.52 & 24.97 & 57.67 & 48.04 & 45.74 & 33.75 & 35.16 & 25.53 \\
    \midrule
    LOCAL-GCL & 51.33 & 40.53 & 47.98 & 36.34 & 54.08 & 27.15 & 61.78 & 48.02 & 46.15 & 33.16 & 38.25 & 29.14 \\
    PHASES & 53.77 & 44.22 & 49.38 & 38.50 & 50.88 & 23.35 & 56.45 & 48.51 & 45.71 & 35.97 & 40.43 & 30.74 \\
    HomoGCL & \underline{56.71} & 47.13 & 49.06 & 38.28 & \underline{58.05} & \underline{33.54} & 62.37 & 53.26 & 48.50 & 37.14 & 40.25 & 31.33 \\
    \midrule
    GRACE+ & 55.36 & 45.92 & 49.91 & 39.62 & 54.20 & 28.36 & 61.88 & \underline{53.94} & 48.08 & 36.66 & \underline{43.59} & \underline{35.02} \\
    ProGCL & 56.00 & 46.52 & 49.90 & 38.98 & 52.14 & 25.60 & 62.14 & 50.32 & 48.27 & 36.44 & 39.12 & 29.21 \\
    GRAPE & 56.50 & \underline{47.45} & \underline{50.87} & \underline{40.39} & OOM & OOM & 62.24 & 53.73 & \underline{48.84} & \underline{38.67} & 40.85 & 32.03 \\
    \midrule
    NML-GCL & \textbf{58.23} & \textbf{49.50} & \textbf{52.46} & \textbf{42.19} & \textbf{59.81} & \textbf{35.82} & \textbf{63.13} & \textbf{55.56} & \textbf{50.52} & \textbf{39.42} & \textbf{45.46} & \textbf{37.74} \\
    \bottomrule
    \end{tabular}%
  }
  \caption{Node clustering results evaluated by FMI (\%) and ARI (\%). 'OOM' means out of memory. The best result is in bold, and the second best is underlined.}
  \label{tab:nodecluster}%
\end{table*}%

\subsection{Performance on Downstream Tasks (RQ1)}
\paragraph{Results of Node Classification.} To train the classifier, we follow the public splits on Cora, CiteSeer, and PubMed, and a 1:1:8 training/validation/testing splits on the other datasets.
The reported results are averaged over 10 runs with random seeds, and the average classification accuracies with standard deviation are reported in Table \ref{tab:nodeclass}. 
We see that NML-GCL consistently outperforms all the baseline methods, especially the soft-weight based ones (GRACE+, ProGCL, and GRAPE) and the hard-weight based ones (LOCAL-GCL, PHASES, and HomoGCL).
Moreover, we conduct the paired t-test on NML-GCL and the best baseline. The p-value in the third-to-last line shows that all the p-values are smaller than 0.01, indicating that the improvement achieved by NML-GCL is statistically  significant.
This improvement is because via the bi-level optimization, NML learns a negative metric space where false negatives are closer to anchor than true negatives. Due to NML, NML-GCL is able to suppress the impact of false negatives according to the learned distance to anchor nodes, which makes the optimization orientation of the encoder be rectified towards better distinguishing false negatives from true negatives, resulting in node embeddings with stronger discriminability. 
In addition, methods addressing false negatives are generally superior to those that do not consider them, highlighting that suppressing the disturbance of false negatives enhances GCL's ability to generate more robust and generalizable node embeddings.

\paragraph{Results of Node Clustering.}
Table \ref{tab:nodecluster} shows FMI and ARI of clustering conducted with $k$-means algorithm over the node embeddings generated by baseline methods and NML-GCL. At first, NML-GCL consistently outperforms the baseline methods in terms of both metrics, indicating that the clustering based on NML-GCL can result in purer clusters, i.e., a cluster contains only nodes of the same label, and the nodes of the same class are grouped into the same cluster. 

\begin{figure}[t]
 \centering
    \subfloat[GRACE]{\includegraphics[width=.45\columnwidth]{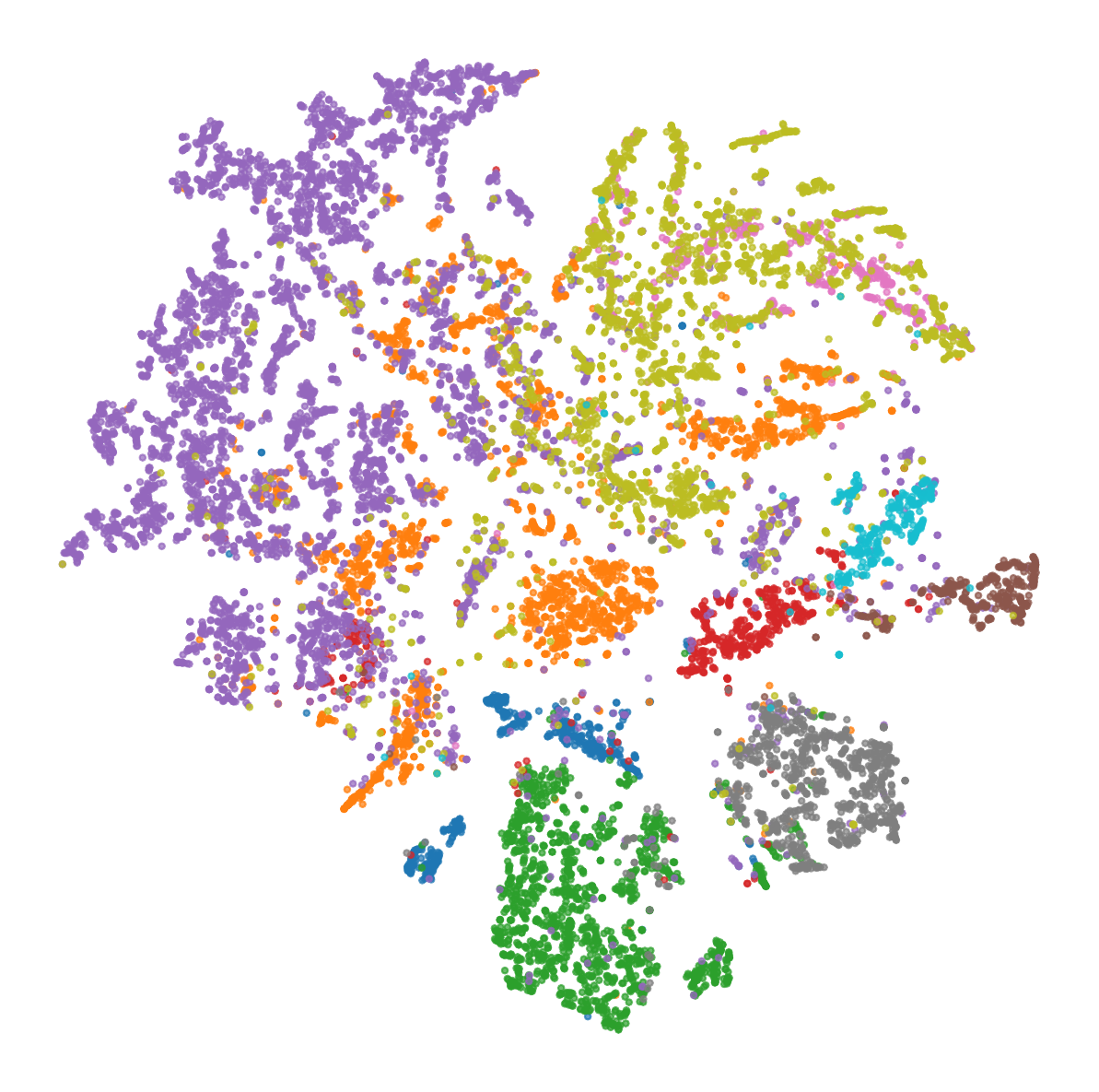}\label{fig:tsne_grace}}\hfil   
    \subfloat[NML-GCL]{\includegraphics[width=.45\columnwidth]{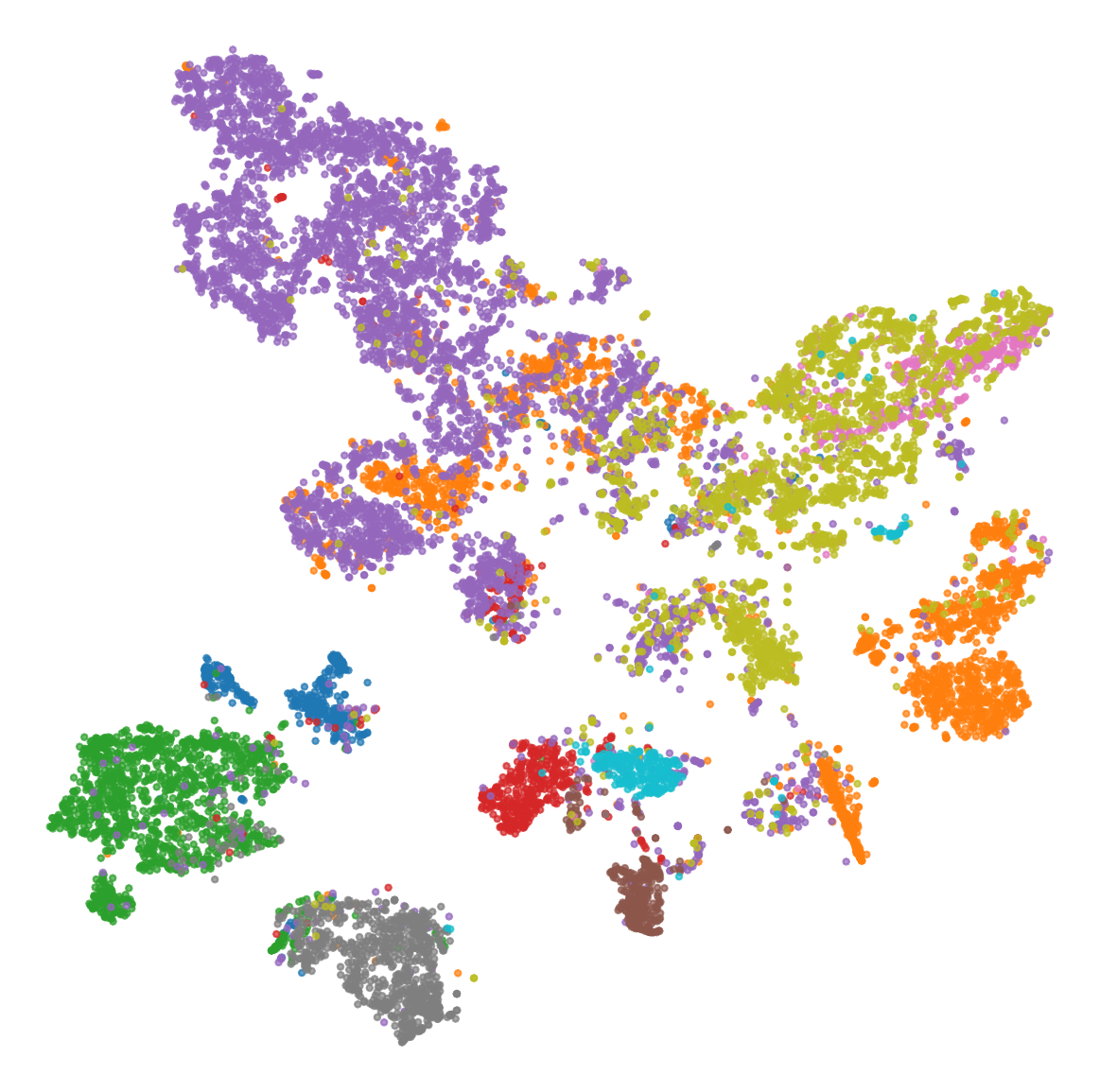}\label{fig:tsne_ours}}
	\caption{Visualization of node clustering on Computers.}
 \label{fig:tsne_computers}
\end{figure}

As a study, we visualize the clustering results of NML-GCL and GRACE on the Computers dataset by t-SNE in Figs. \ref{fig:tsne_grace} and \ref{fig:tsne_ours}, respectively. We can see that compared with the traditional InfoNCE based method GRACE, NML-GCL can produce clustering results with higher intra-cluster cohesion and clearer inter-cluster boundaries. This is because NML-GCL, aided by negative metric learning, brings positive examples (including false negatives) closer together and pushes positive and true negative examples farther apart more effectively.

\subsection{Ablation Study (RQ2)}
In this section, we demonstrate the necessity of NMN $M$ in NML-GCL. Specifically, we compare NML-GCL with the following variants: (1) \textbf{w/o $M$:} We remove the negative metric network, and $m_{ij}$ is a constant (i.e., $\frac{1}{N}$); (2) \textbf{repl. cosine sim.:} $m_{ij}$ is calculated as $exp(-cosine\_similarity(u_i,v_i))$. 

As shown in the last two rows of Table \ref{tab:nodeclass}, we see the NMN component contributes to the performance improvement of NML-GCL, especially with a 3\% gain on CiteSeer. Moreover, although the use of cosine similarity can alleviate the problem of false negatives to some extent, NML-GCL uses the NMN $M$ to capture the nonlinearity in the distance measurement, so that false negatives can be better distinguish from true negatives in the negative metric space. Overall, the results highlight that the inclusion of $M$ in NML-GCL is essential for refining the negative sample learning process, enabling a more nuanced and effective representation of negative pairwise distances. This demonstrates the critical role of NMN in achieving superior performance compared to simpler, predefined negative sample weighting strategies like cosine similarity.

\subsection{Hyper-parameter Analysis (RQ3)}
\begin{figure}[t]
 \centering
        \subfloat[$\alpha$]{
	\includegraphics[width=.48\columnwidth]{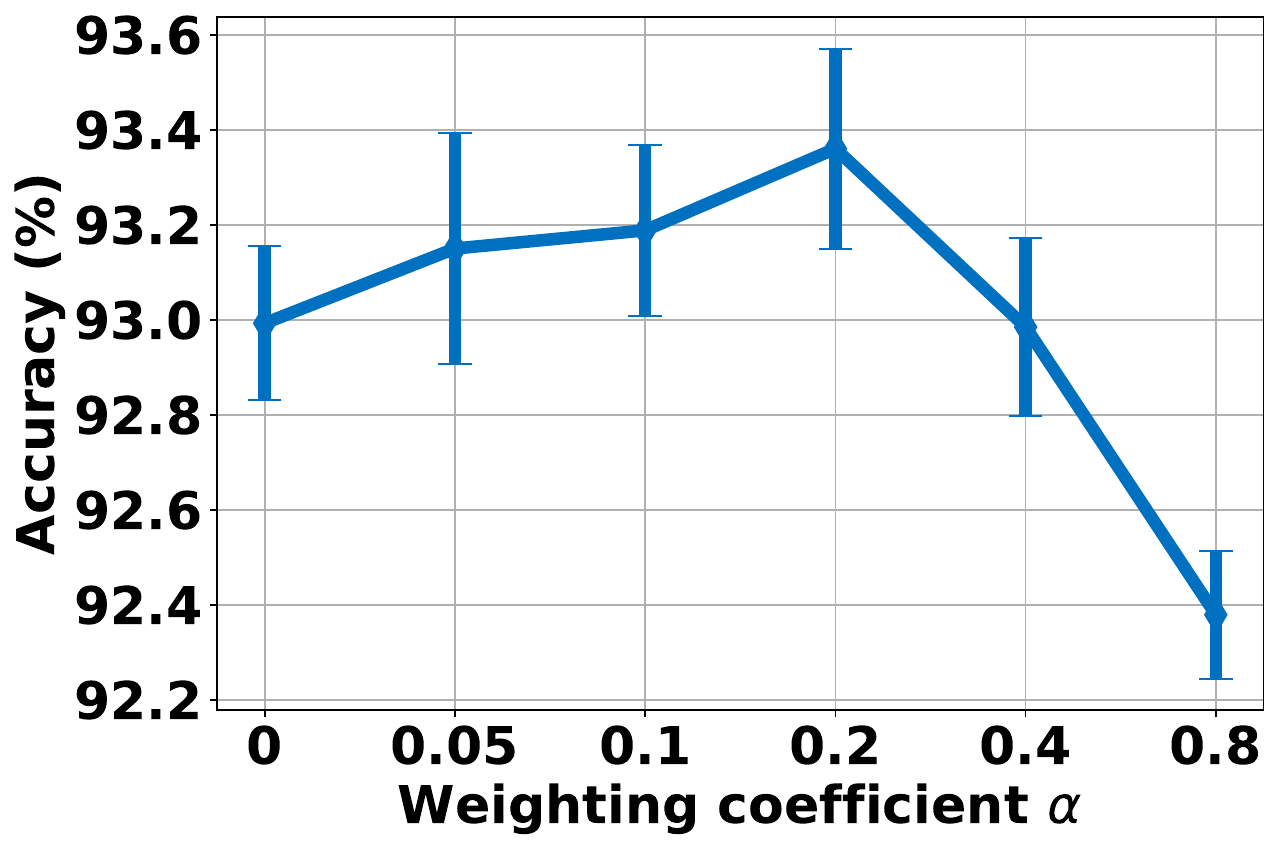}\label{fig:alpha}}
        \subfloat[$T_\text{M}$]{
	\includegraphics[width=.48\columnwidth]{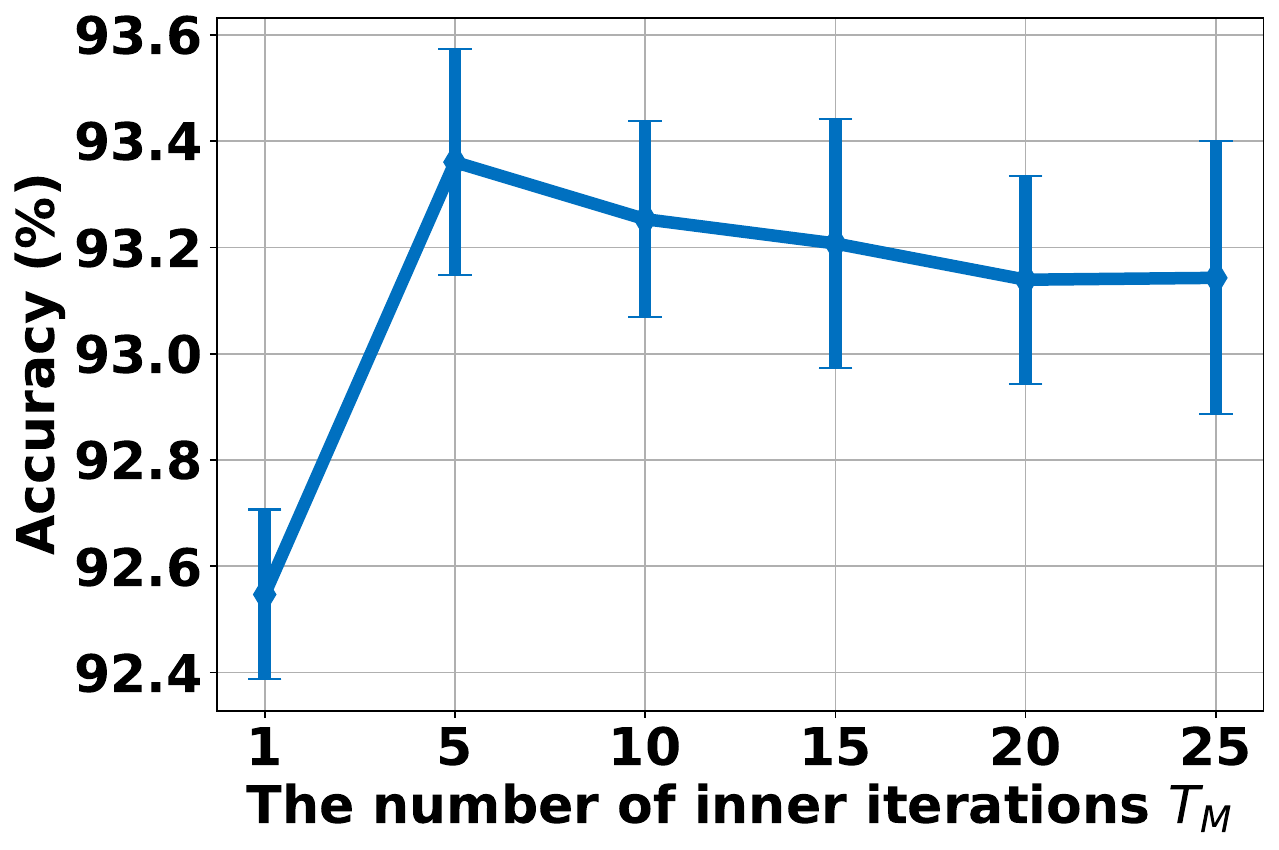}\label{fig:tm}}

	\caption{Hyper-parameter analysis. }
 \label{fig:param}
\end{figure}

Now we investigate the effectiveness of the two most important hyper-parameters $\alpha$ in Equation (\ref{eq:objective}) and the iteration number $T_{\text{M}}$ of inner minimization in Algorithm \ref{algorithm}. We take the Photo dataset as an example. Fig. \ref{fig:alpha} shows that as $\alpha$ increases, the classification accuracy first rises and then falls. Since $\alpha$ is the weight of the regularization term in Equation (\ref{eq:objective}), this result indicates that when $\alpha$ is small, NML-GCL suffers from overfitting. In contrast, when $\alpha$ becomes large, the weight distribution of negatives tends to become uniform, causing the learned embeddings to lose their discriminability. 
From Fig. \ref{fig:tm}, we see that too few iterations of the inner minimization can lead to underfitting, while too many iterations can cause overfitting and unnecessary wasting of time.
% We have similar observation from Fig. \ref{fig:tau}. From Fig. \ref{fig:tau}, we see that  both excessively large and small values of $\tau$ reduce accuracy. This is because a very small $\tau$ causes the learned weights to degrade into hard weights of either 0 or 1, while a very large $\tau$ results in the weights converging toward similar values. Both cases diminish the discriminability of the learned embeddings. 
% From Fig. \ref{fig:tm}, we see that too few iterations of the inner minimization will lead to underfitting, whereas too many iterations can cause overfitting.

\section{CONCLUSION}\label{conclusion}
In this paper, we introduce a novel approach called NML-GCL. NML-GCL utilizes a learnable negative metric network to construct a negative metric space, allowing for better differentiation between false negatives and true negatives based on their distances to the anchor node. To address the challenge of the lack of explicit supervision signals for NML, we present a joint training scheme with a bi-level optimization objective that implicitly leverages self-supervision signals to iteratively refine both the encoder and the negative metric network. Comprehensive theoretical analysis and extensive experiments conducted on widely used benchmarks demonstrate the superiority of our proposed method to baseline methods on downstream tasks.

\clearpage
\section*{Acknowledgements}
This work is supported by Natural Science Foundation of Sichuan Province under grant 2024NSFSC0516 and National Natural Science Foundation of China under grant 61972270.

\bibliographystyle{named}
\bibliography{refs}

\clearpage
\appendix

\section{RELATED WORK}\label{related_work}
In the GCL domain, the existing methods typically address the issue of false negatives by weighting negative samples, following two technique lines: hard-weight based \cite{zhang2022local,wang2024select,hu2021graph,liu2023b2,wuconditional,huynh2022boosting,fan2023neighborhood,han2023topology,sun2023progressive,yang2022region,li2023homogcl,liu2024seeking} and soft-weight based \cite{xia2022progcl,lin2022prototypical,hao2024towards,liu2024seeking,niu2024affinity,zhuo2024improving,wan2023boosting,chi2024enhancing,zhuo2024graph}. Hard-weight methods typically assign binary weights to negative samples using predefined criteria like similarity thresholds \cite{wuconditional,huynh2022boosting} or neighborhood distances \cite{li2023homogcl,liu2024seeking,zhang2022local}. For example, LOCAL-GCL \cite{zhang2022local}, and HomoGCL \cite{li2023homogcl} identify false negatives based on first-order neighbors, while PHASES \cite{sun2023progressive} uses a feature similarity threshold. Soft-weight based methods relax the weight to $[0,1]$. For example, GRACE+ \cite{chi2024enhancing} determines the weights of negatives using a predefined negative sample distribution. Some soft-weight based methods determine the weights by clustering. For example, ProGCL \cite{xia2022progcl} and GRAPE \cite{hao2024towards} employ a two-component beta mixture model and a self-expressive learning objective, respectively, to identify false negatives.

However, both hard-weight-based and soft-weight-based methods rely on human prior knowledge (e.g., the threshold or the number of clusters), leading to suboptimal results. In contrast, NML-GCL is a data-driven approach that can automatically learn the weights of negatives from data through the novel Negative Metric Learning.

\begin{figure*}[t]
 \centering
 \subfloat[Cora]{\includegraphics[width=.15\textwidth]{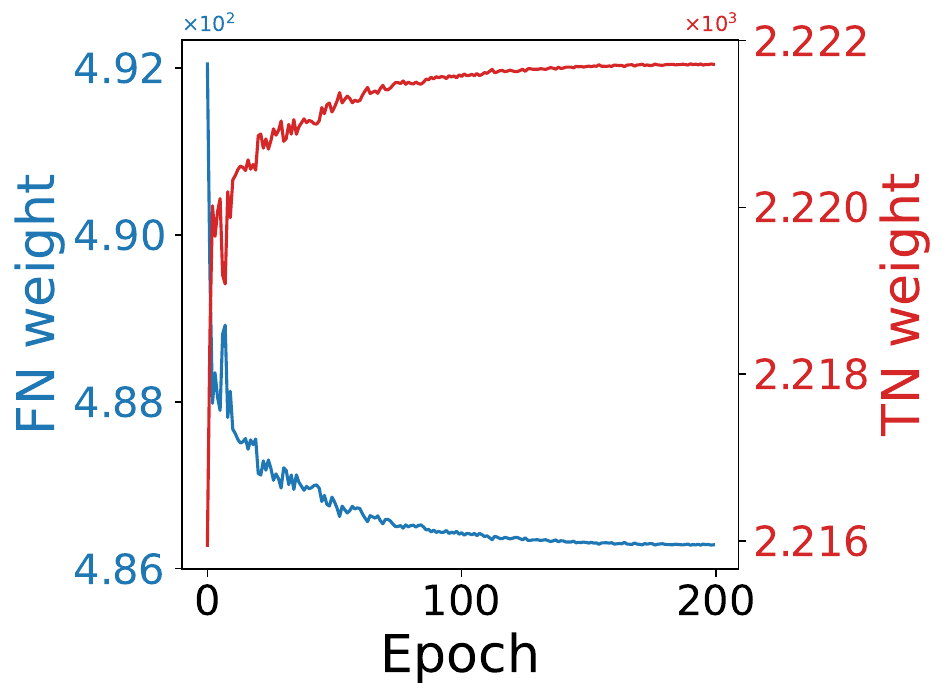}}\hfil
 \subfloat[CiteSeer]{\includegraphics[width=.15\textwidth]{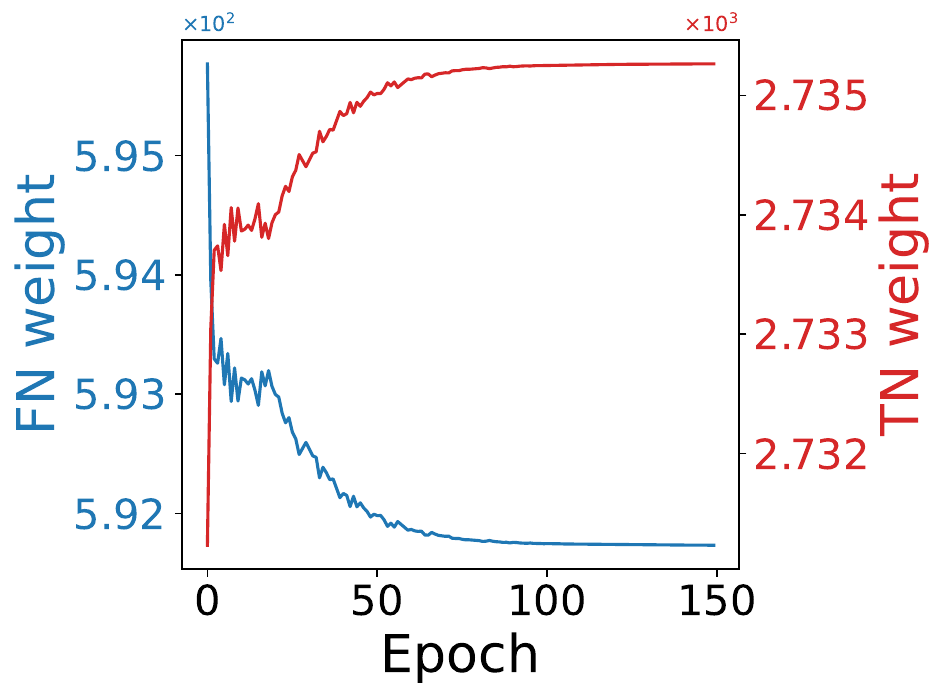}}\hfil
 \subfloat[PubMed]{\includegraphics[width=.15\textwidth]{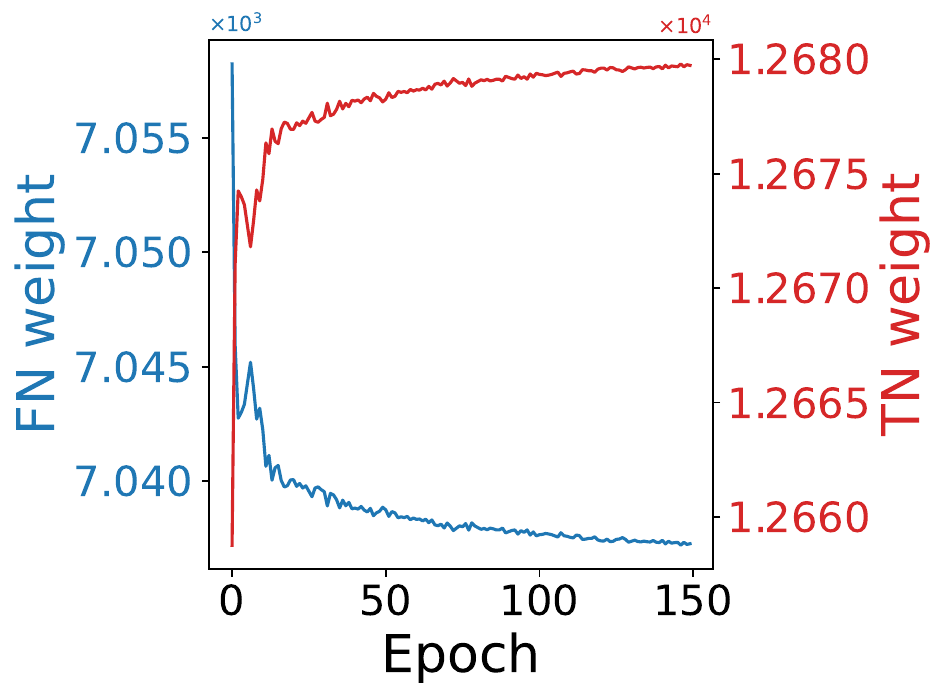}}\hfil
 \subfloat[Photo]{\includegraphics[width=.15\textwidth]{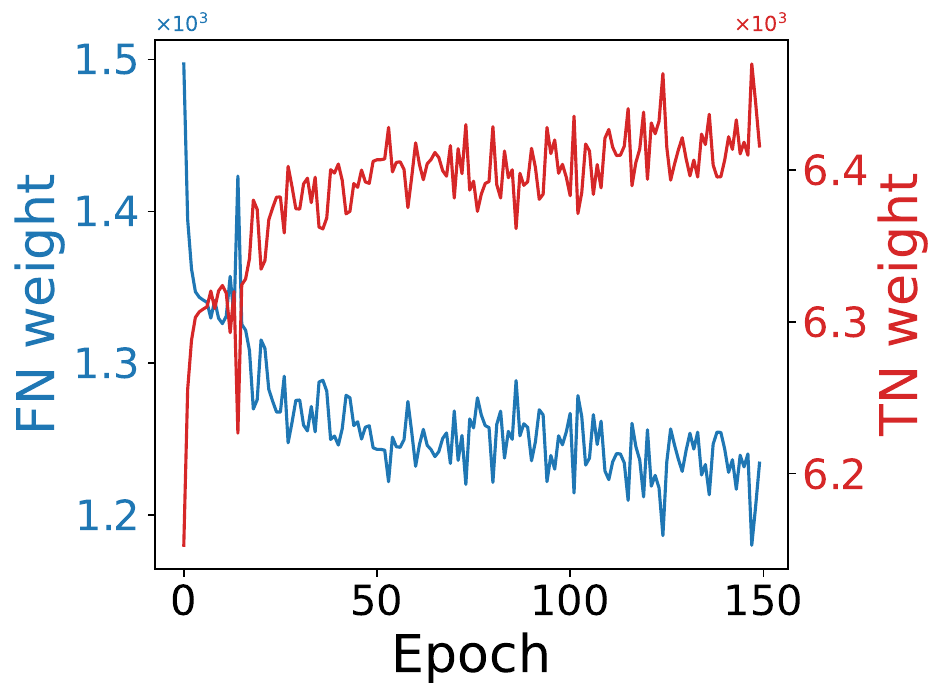}}\hfil
 \subfloat[Computers]{\includegraphics[width=.15\textwidth]{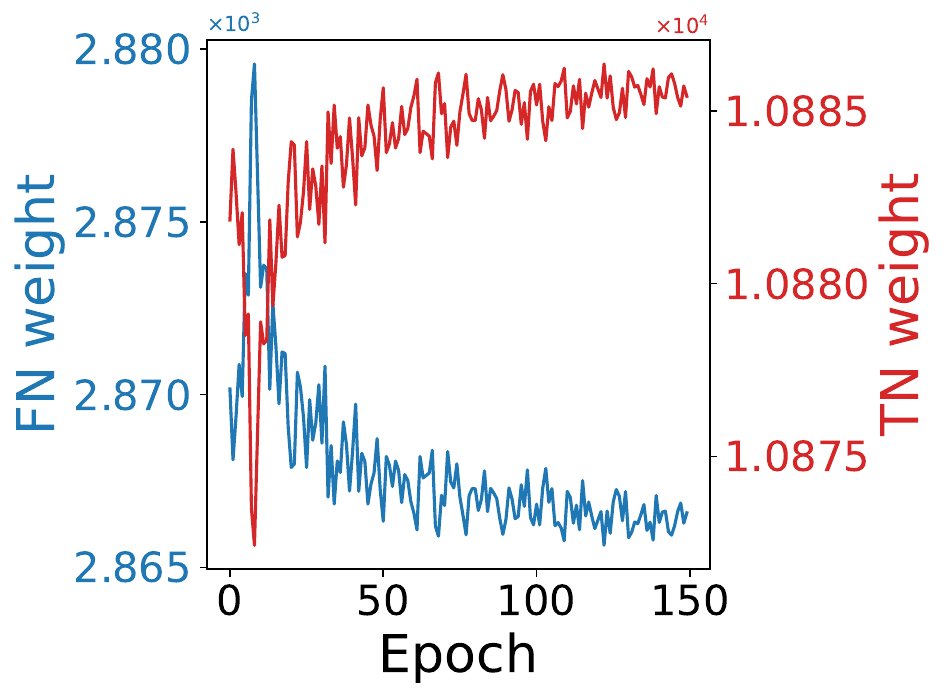}}\hfil
 \subfloat[Wiki-CS]{\includegraphics[width=.15\textwidth]{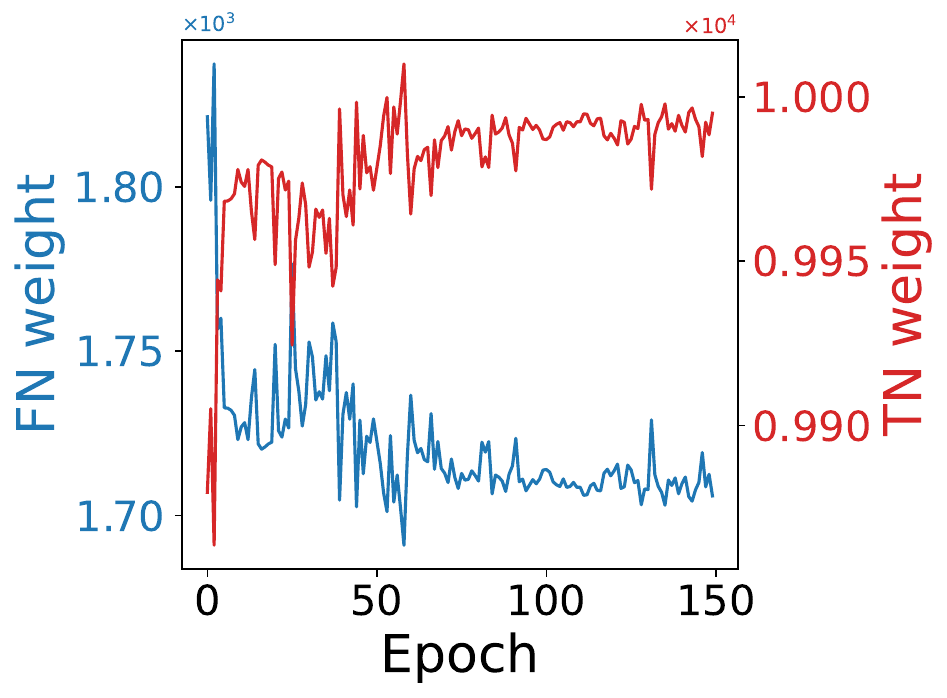}}
 
 \caption{Predicted weights of False Negatives (FN) and True Negatives (TN) during training of NML-GCL.}
 \label{fig:fn_tn_weights}
\end{figure*}

\begin{figure*}[t]
 \centering
 \subfloat[Cora original feature]{\includegraphics[width=.15\textwidth]{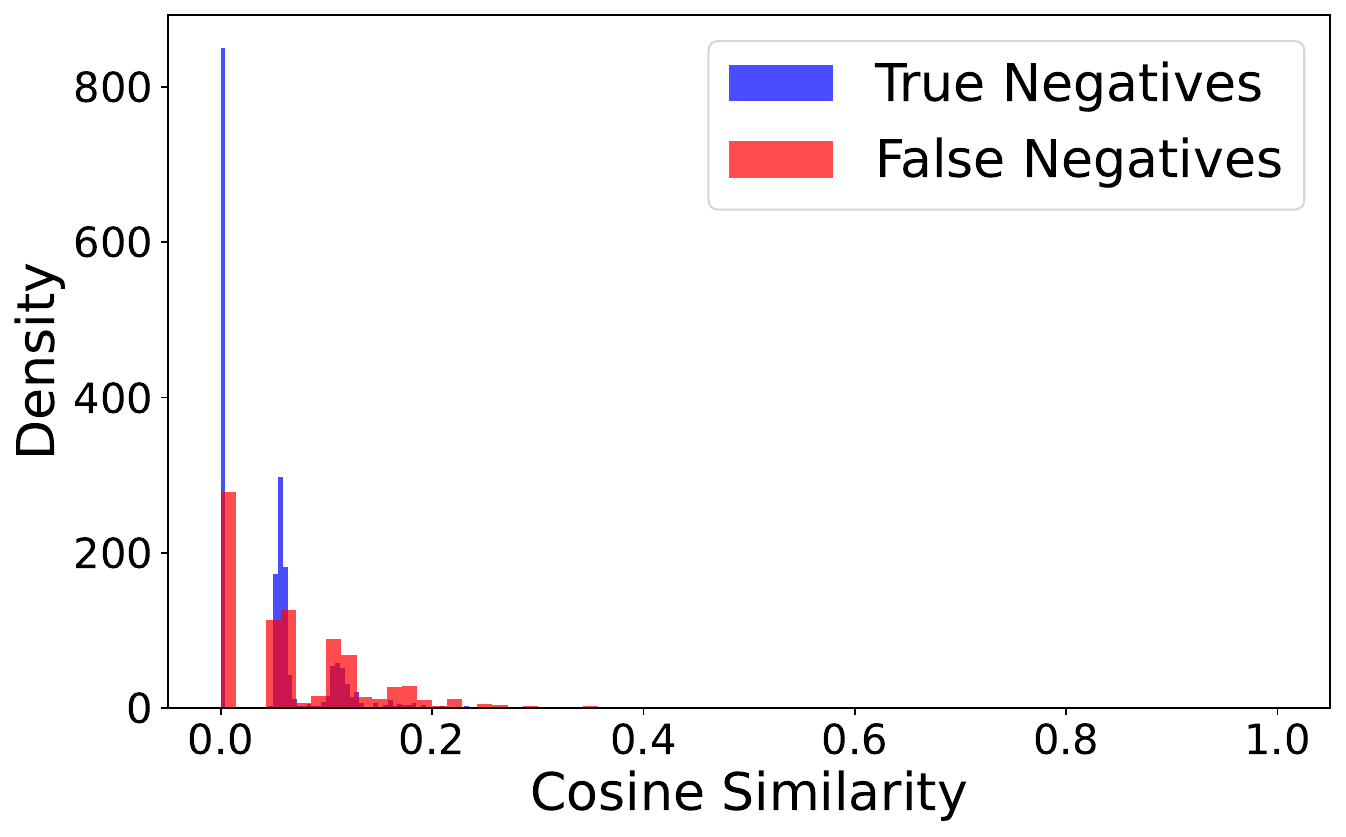}\label{fig:negative_dist_orig_cora}}\hfil
\subfloat[CiteSeer original feature]{\includegraphics[width=.15\textwidth]{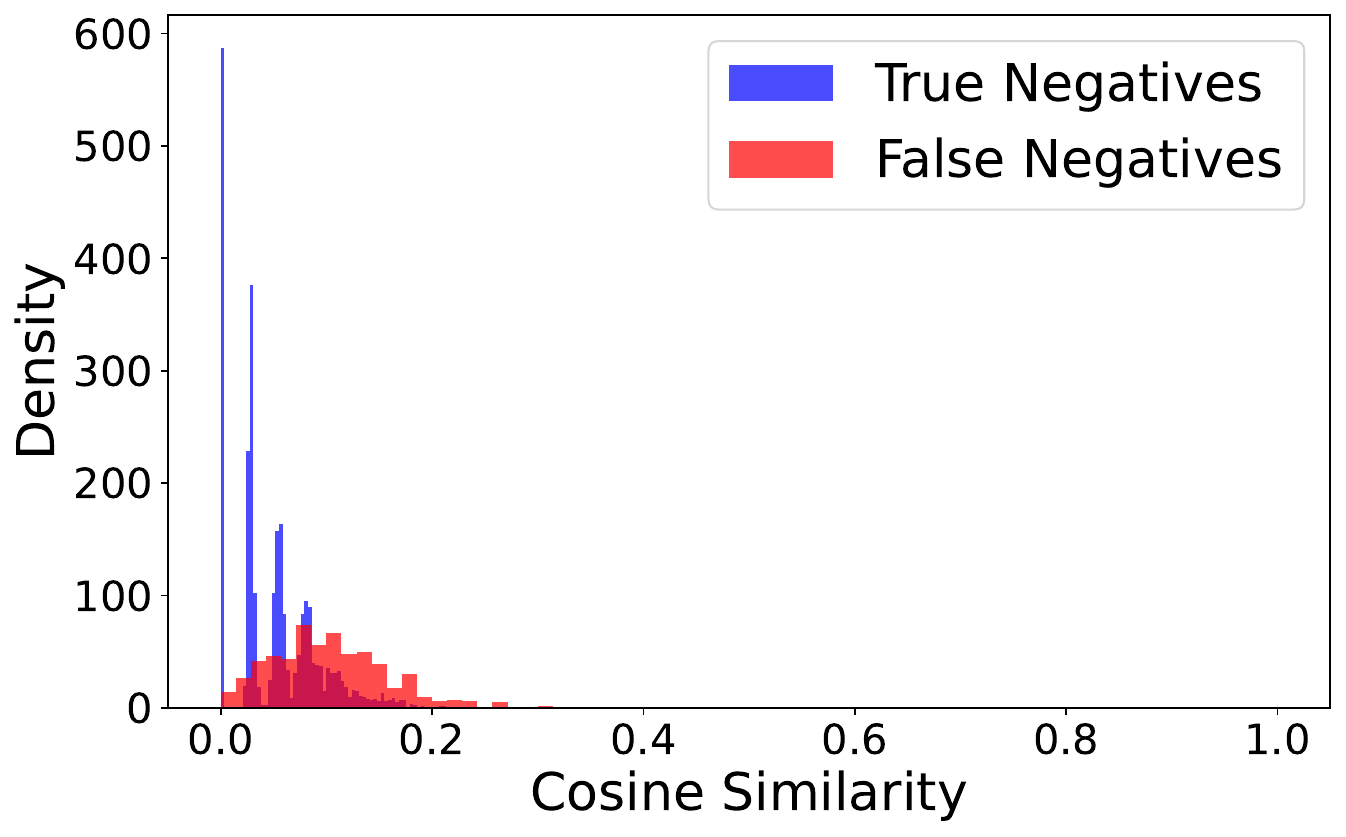}\label{fig:negative_dist_orig_citeseer}}\hfil
\subfloat[PubMed original feature]{\includegraphics[width=.15\textwidth]{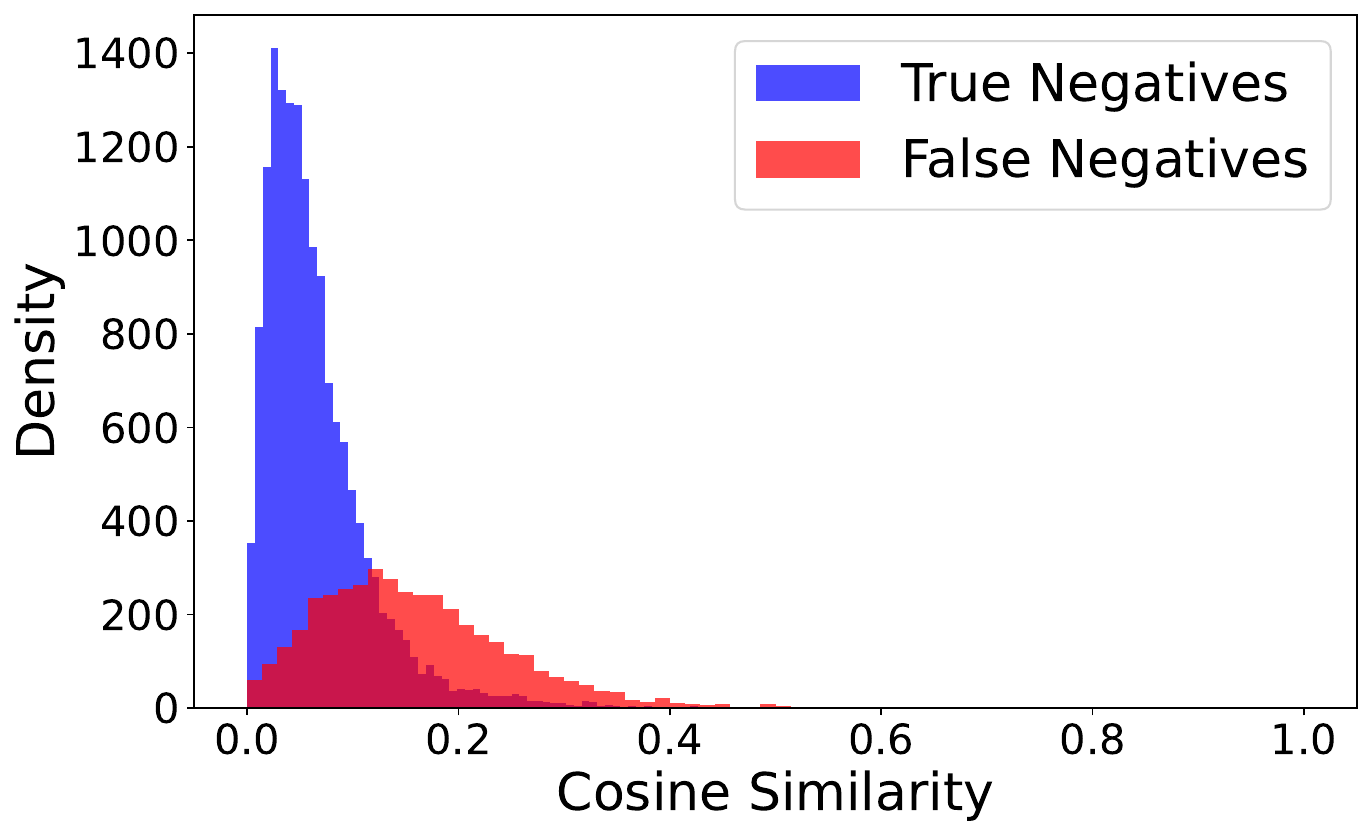}}\hfil
\subfloat[Photo original features]{\includegraphics[width=.15\textwidth]{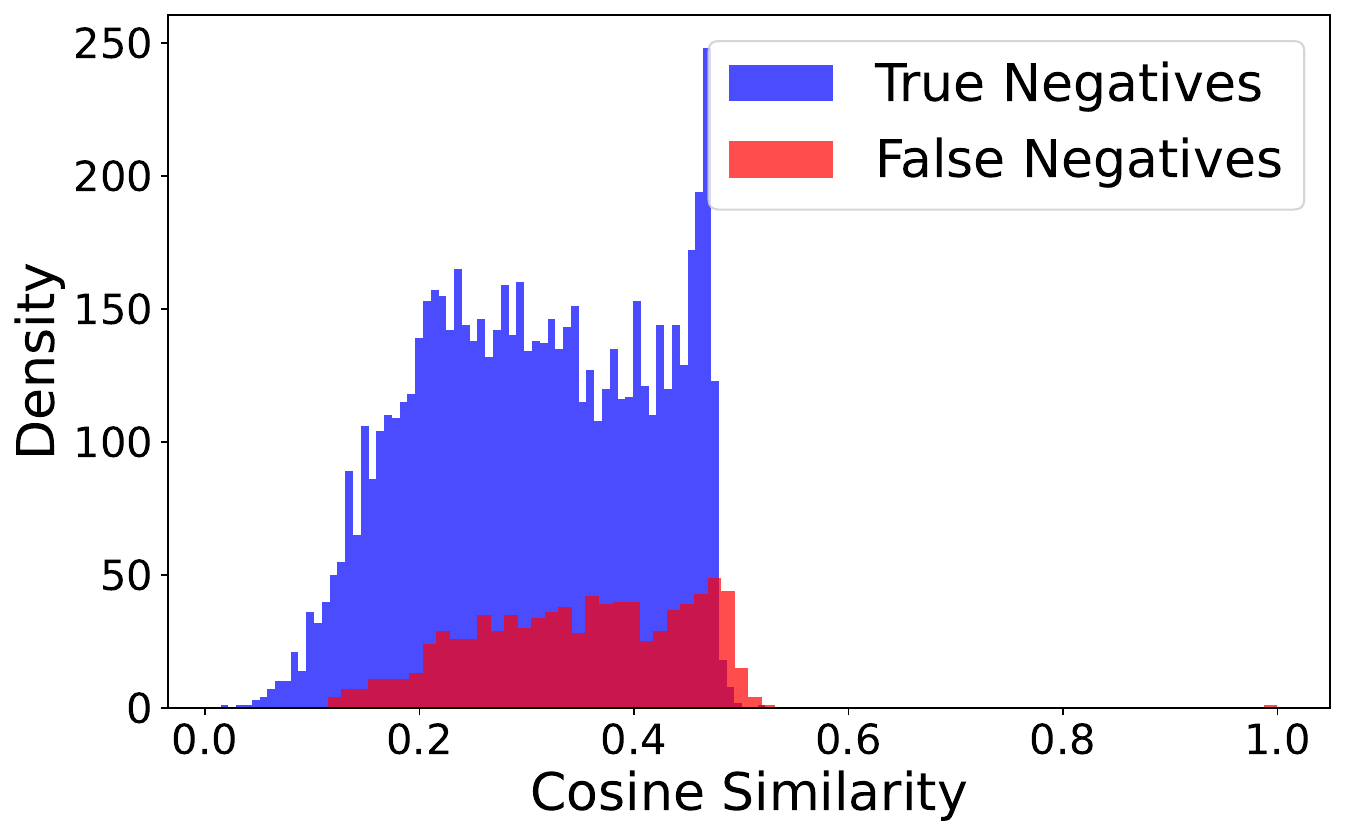}}\hfil
 \subfloat[Computers original features]{\includegraphics[width=.15\textwidth]{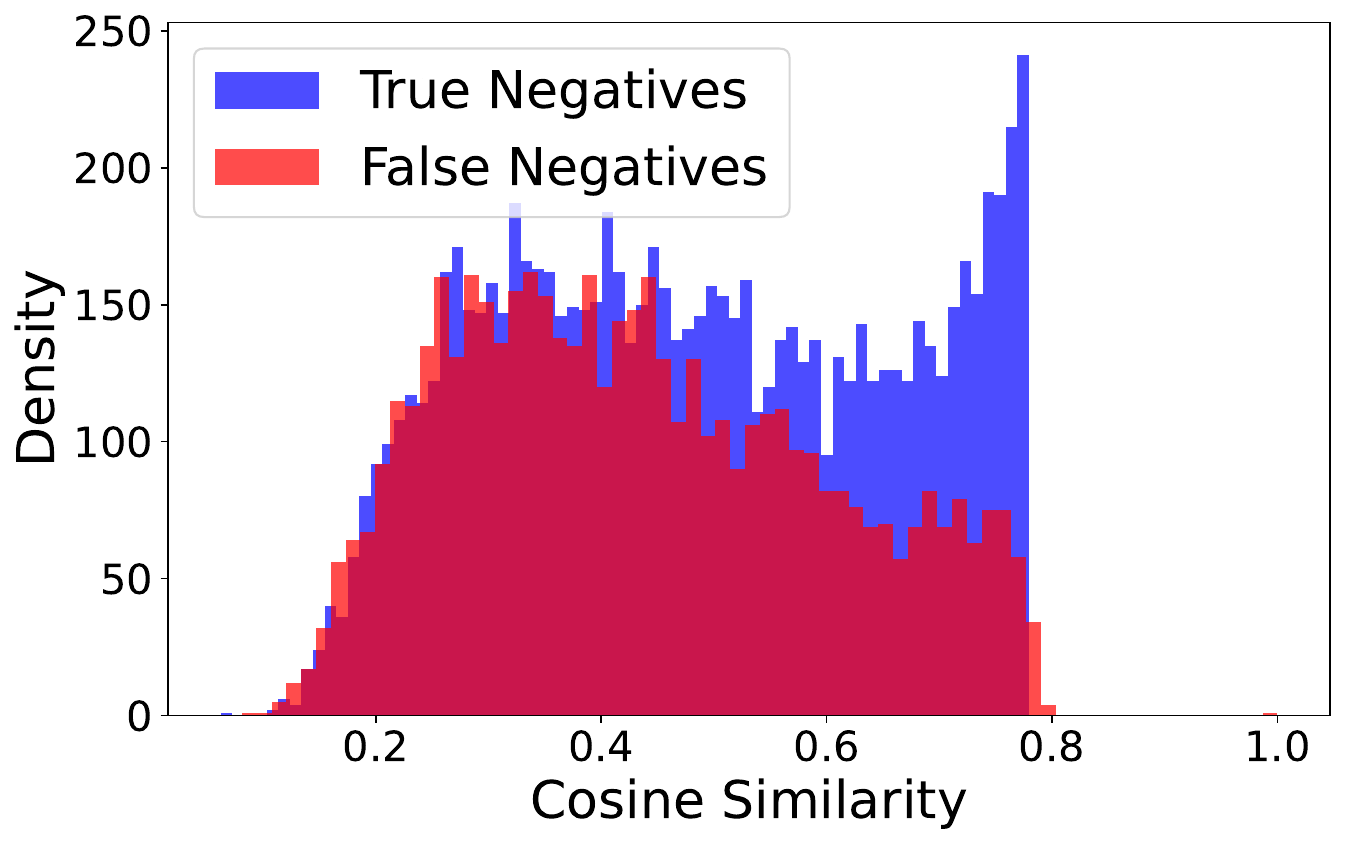}}\hfil
 \subfloat[Wiki-CS original feature]{\includegraphics[width=.15\textwidth]{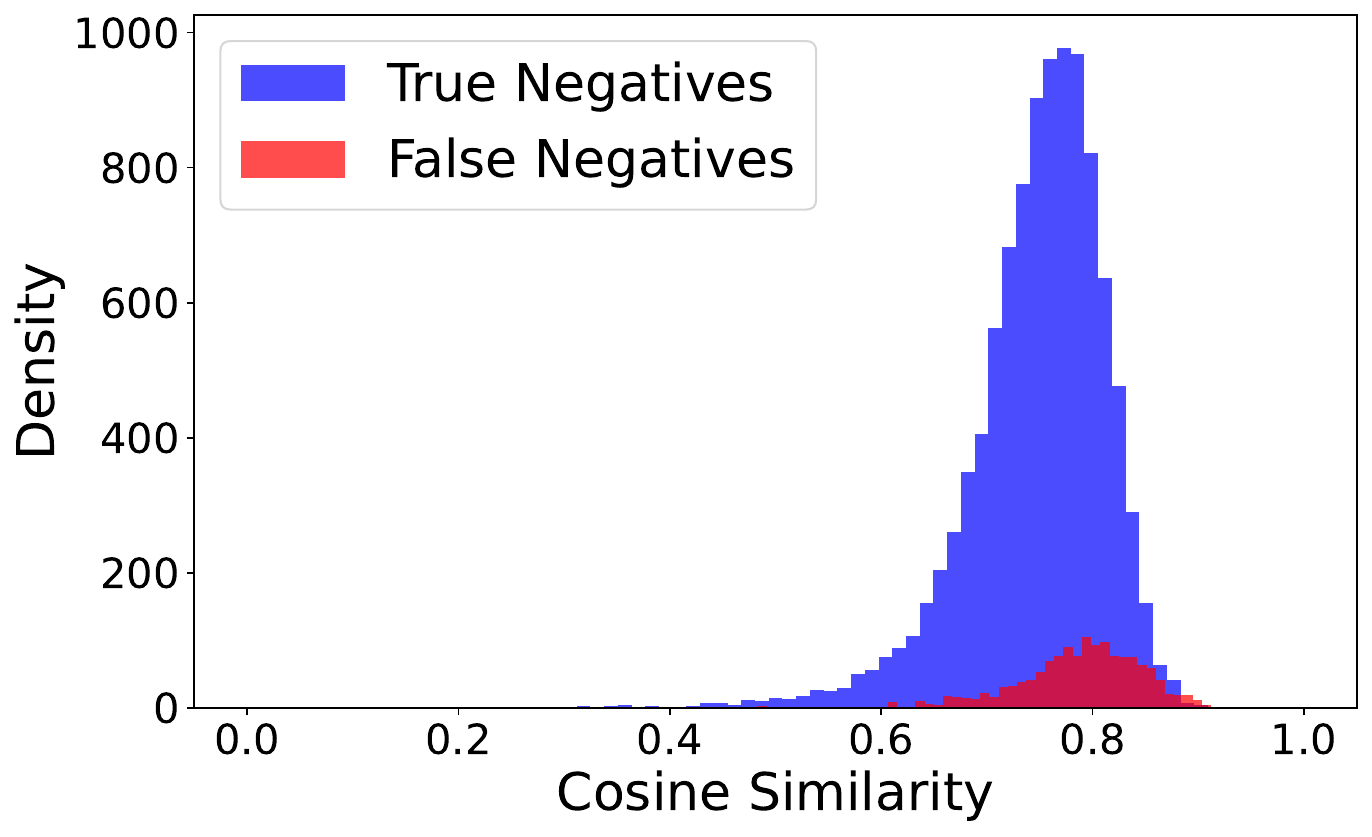}}

 \subfloat[Cora embeddings]{\includegraphics[width=.15\textwidth]{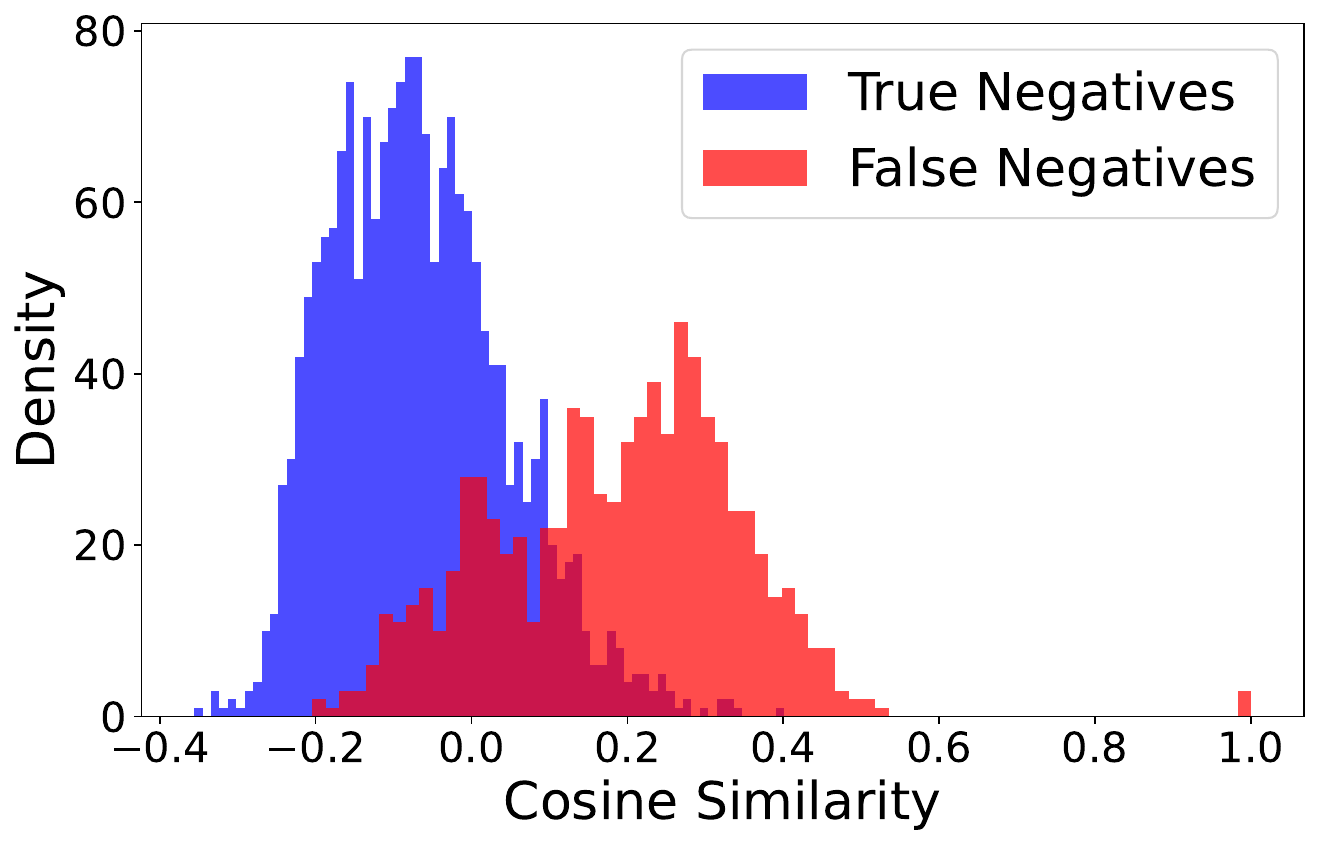}\label{fig:negative_dist_nmlgcl_cora}}\hfil
 \subfloat[CiteSeer embeddings]{\includegraphics[width=.15\textwidth]{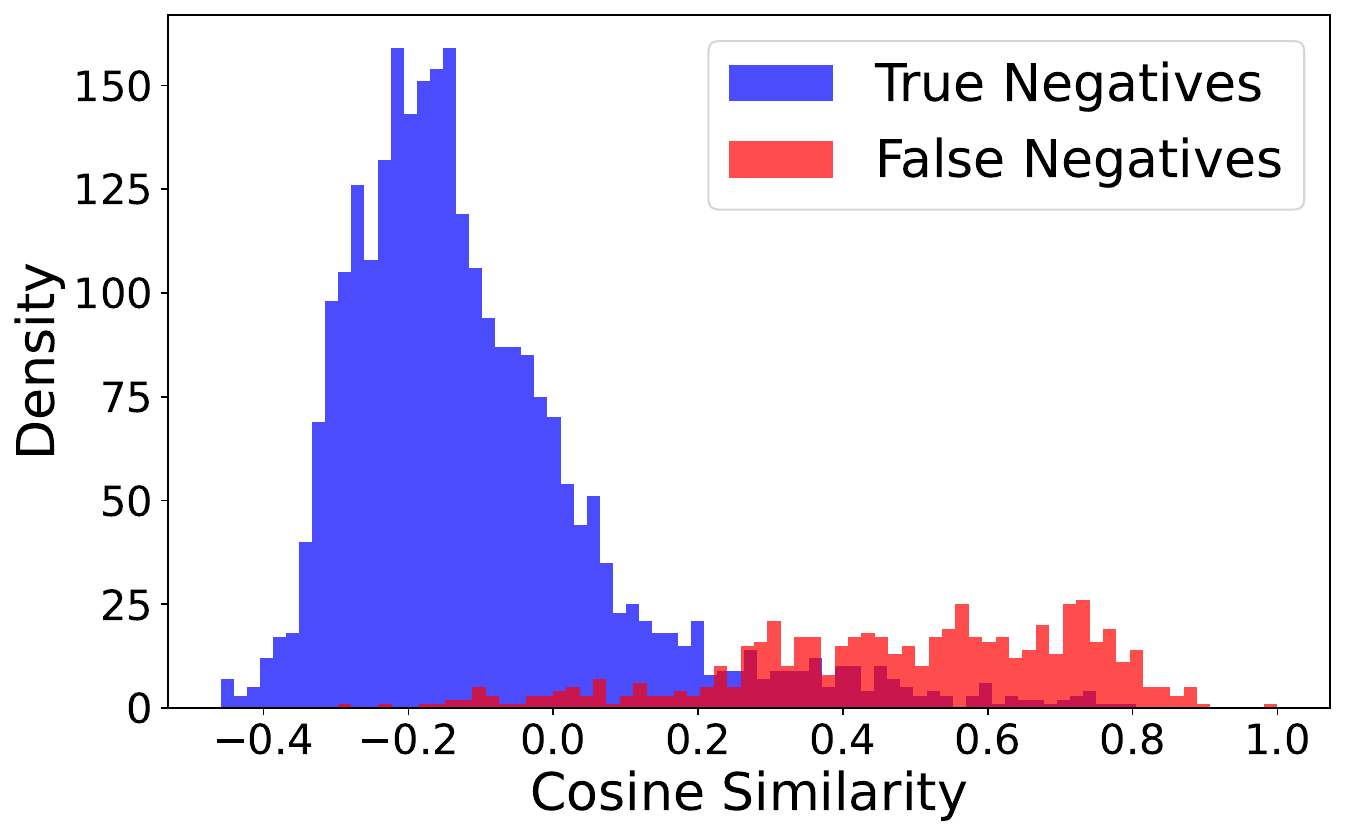}\label{fig:negative_dist_nmlgcl_citeseer}}\hfil
 \subfloat[PubMed embedding]{\includegraphics[width=.15\textwidth]{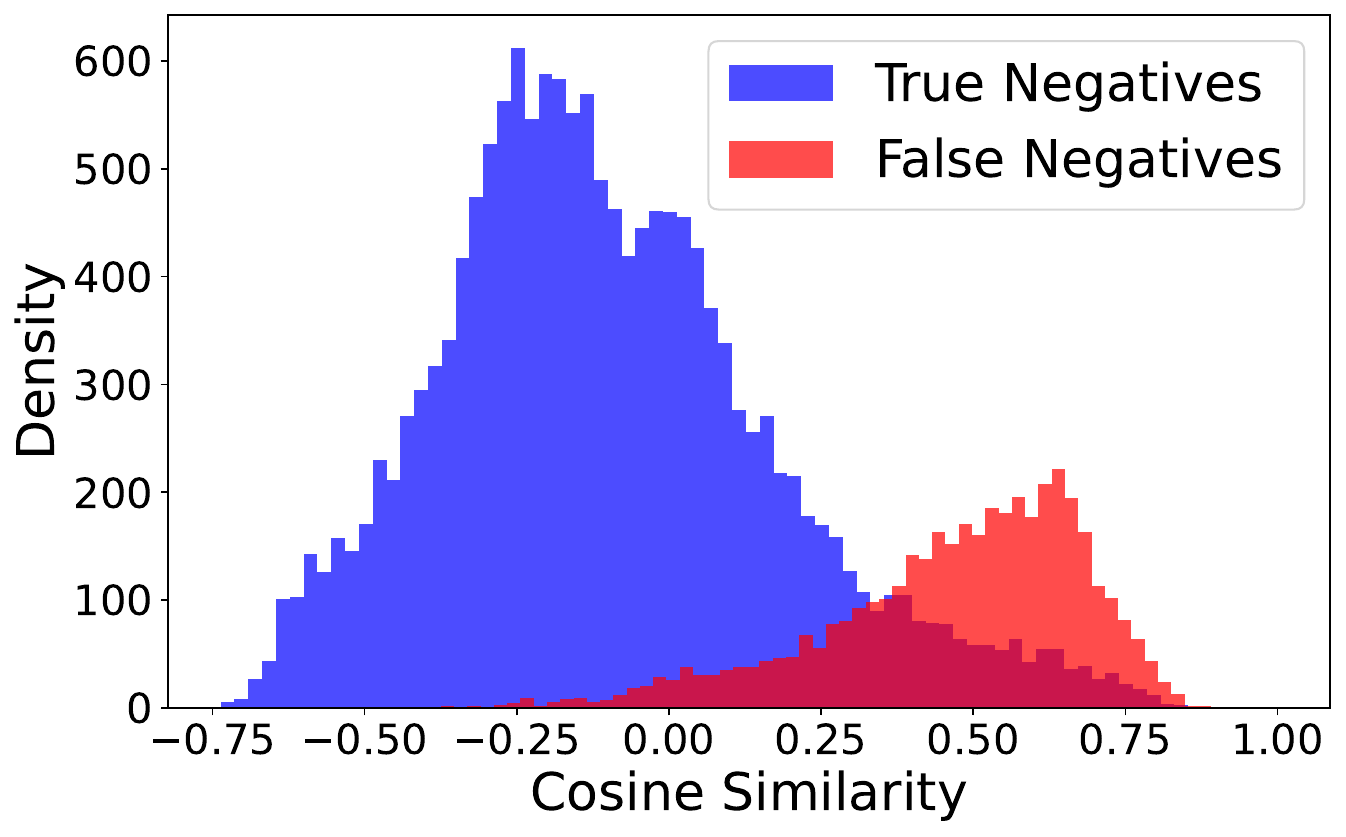}}\hfil
 \subfloat[Photo embeddings]{\includegraphics[width=.15\textwidth]{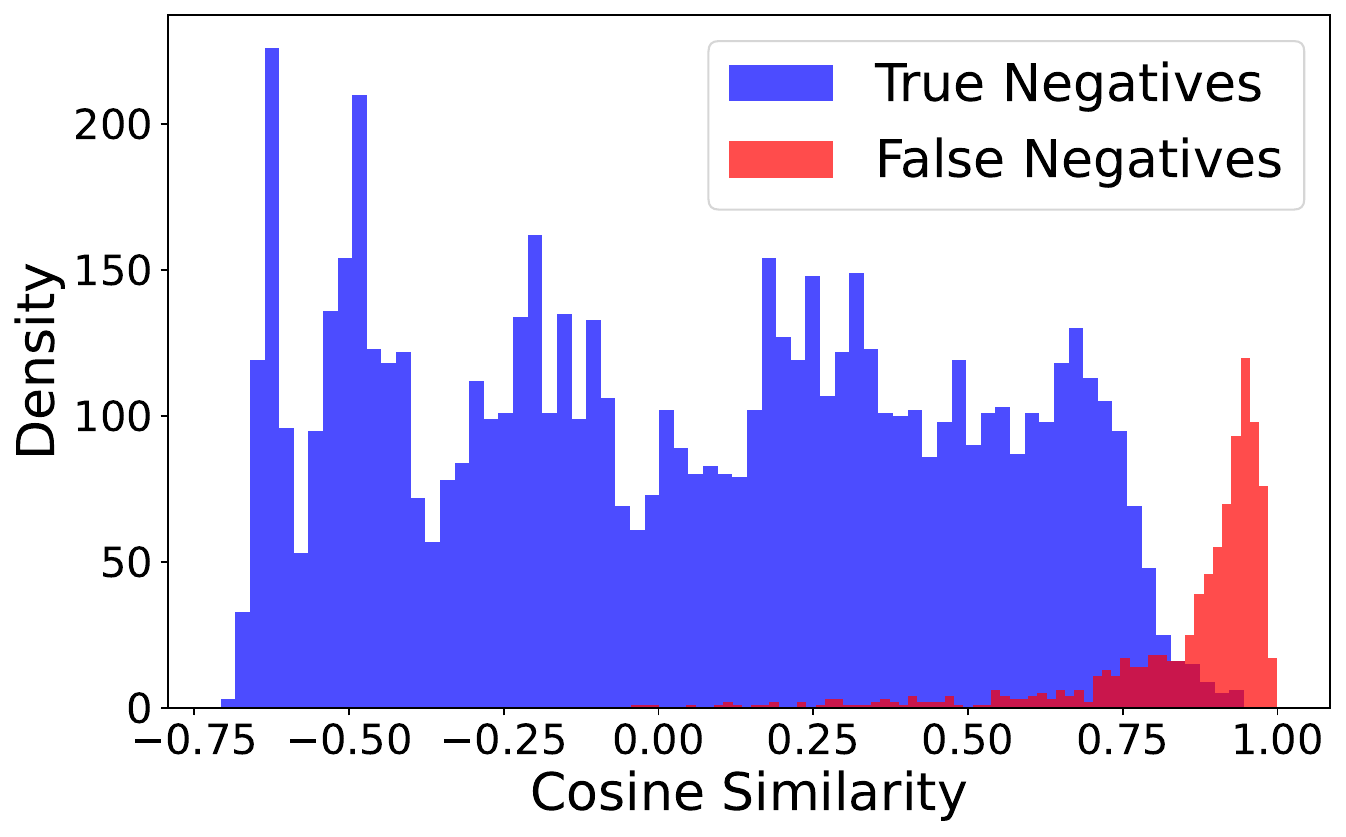}}\hfil
 \subfloat[Computers embeddings]{\includegraphics[width=.15\textwidth]{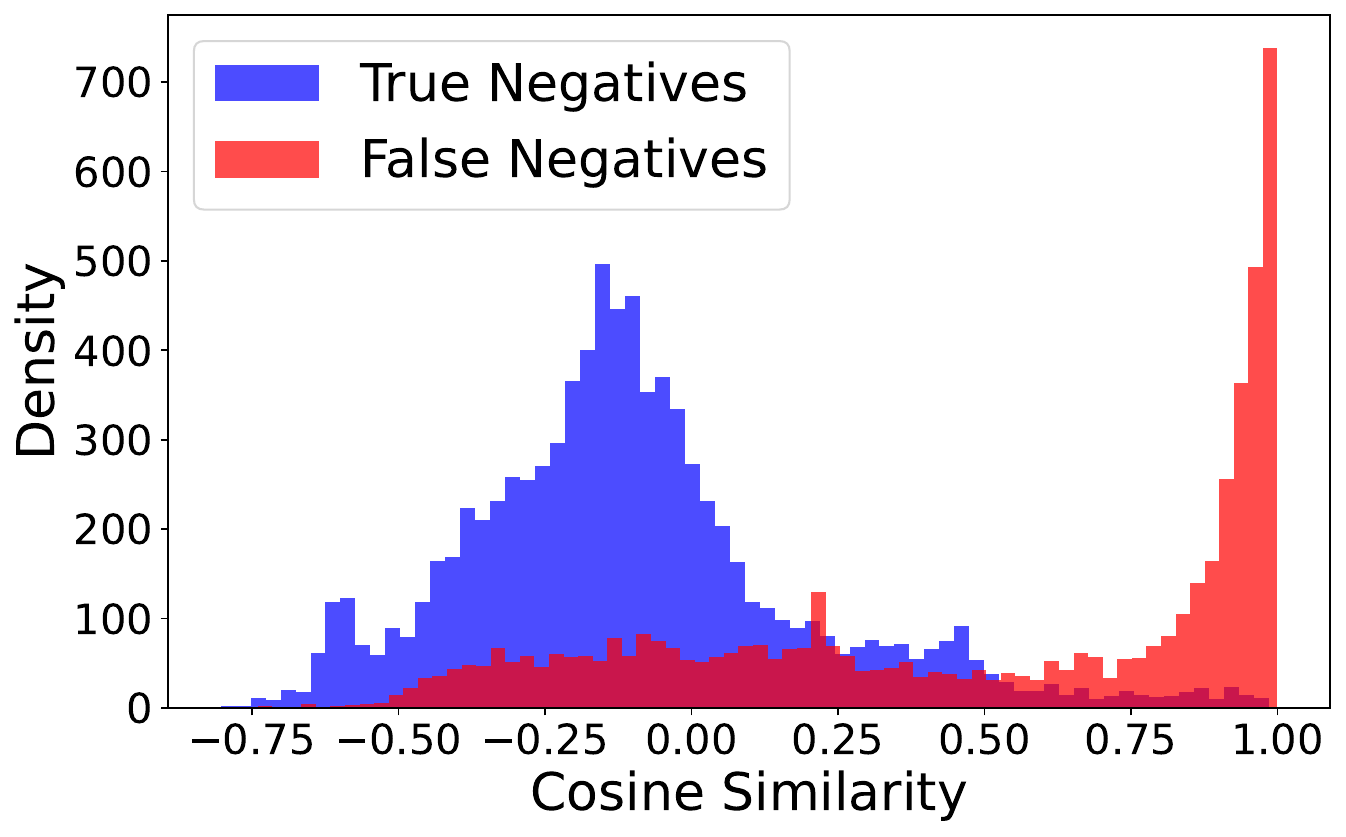}}\hfil
 \subfloat[Wiki-CS embedding]{\includegraphics[width=.15\textwidth]{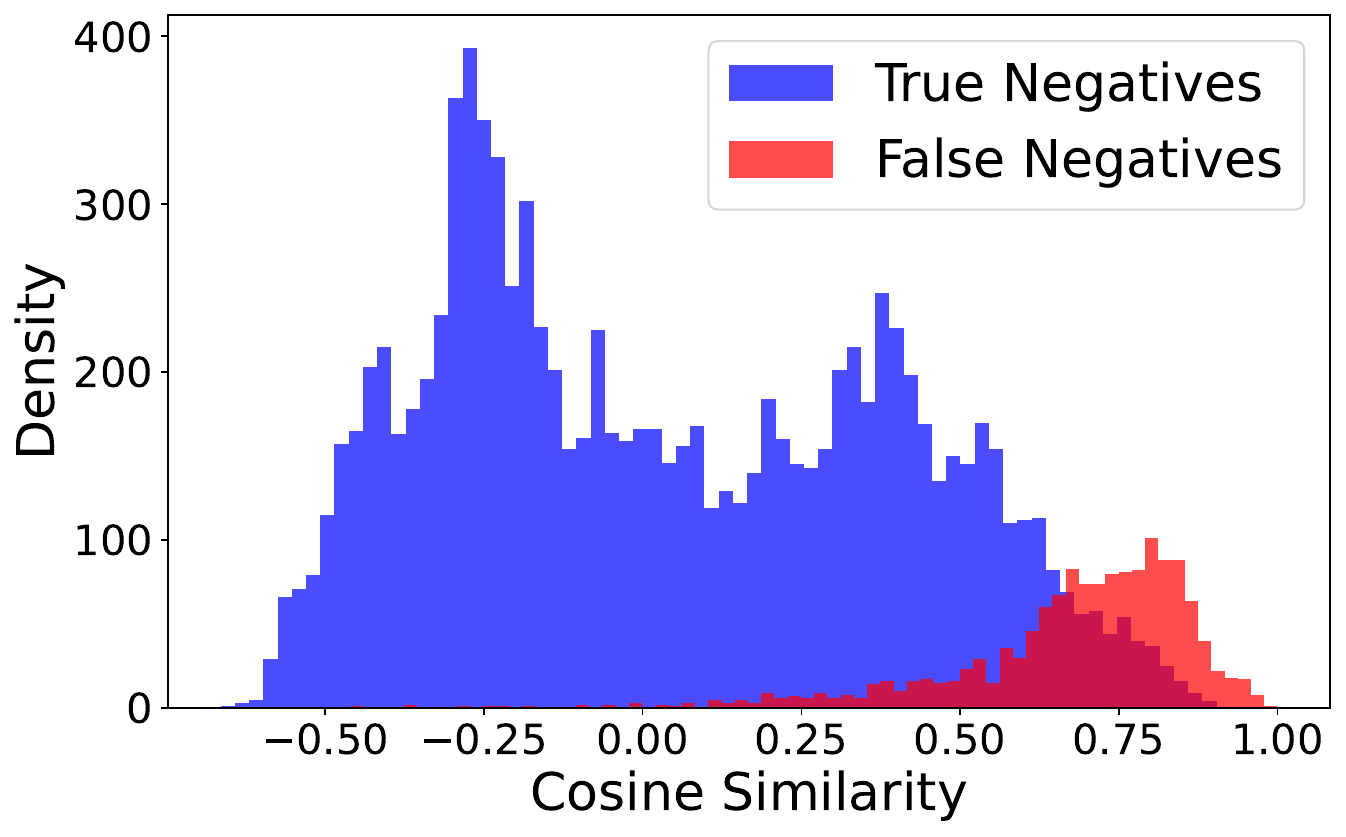}}
 
 \caption{Similarity distributions of negatives with anchors.}
 \label{fig:negative_dist}
\end{figure*}

\section{Proofs}
Without ambiguity, we use $s_{ij}$ to denote $e^{\theta (\mathbf{u}_i,\mathbf{v}_j)}$ for simplicity. 

\subsection{Proof of Theorem \ref{th:tighter}}\label{proof-th1}
\begin{proof}
We first analyze the relationship between $I_{\text{NML}} (U; V)$ and the $I ( U; V )$ to show that $I_{\text{NML}} (U; V)$ is a lower bound of $I ( U; V )$, and then analyze the relationship between $I_{\text{NML}} (U; V)$ and the $I_{\text{NCE}} ( U; V )$ to justify $I_{\text{NML}} (U; V)$ is a tighter bound than InfoNCE. 

\textbf{(1) Lower Bound of MI. }
Let $P_{UV}$ be the joint distribution of the two embedding variables $U$ and $V$, while $P_{U}$ and $P_{V}$ are the marginal distributions. According to the definition of MI \cite{alajaji2018introduction} and Lemma 1 of \cite{nguyen2010estimating}
\begin{small}
\begin{equation*}
\begin{aligned}
& I (U; V) = \text{KL} (P_{UV} \Vert P_{U} P_{V}) \\
= & \max_{f} \big\{ \mathbb{E}_{(\mathbf{u}_i, \mathbf{v}_i) \sim P_{UV}} [f(\mathbf{u}_i, \mathbf{v}_i)] - \mathbb{E}_{(\mathbf{u}_i, \mathbf{v}_i) \sim P_{U} P_{V} } [e^{f(\mathbf{u}_i, \mathbf{v}_i)}] + 1 \big\}, 
\end{aligned}
\end{equation*}
\end{small}
where $f$ is a measurable function $\mathbb{R}^d \times \mathbb{R}^d \mapsto \mathbb{R}$. Now let $f(\mathbf{u}_i, \mathbf{v}_i) = \log \frac{N s_{ii}}{s_{ii} + (N-1) \sum_{j = 1}^{N}\mathbb{E}_{(\mathbf{u}_i, \mathbf{v}_j) \sim P_{U}P_{V}} m_{ij} s_{ij}} $, then we have
\begin{tiny}
\begin{equation*}
\begin{aligned}
& I (U; V) \\
\ge & \max_{ \{m_{ij}\} } \Big\{ \mathbb{E}_{(\mathbf{u}_i, \mathbf{v}_i)\sim P_{UV} } \big[\log \frac{N s_{ii}}{s_{ii} + (N-1) \sum_{j = 1}^{N}\mathbb{E}_{(\mathbf{u}_i, \mathbf{v}_j) \sim P_{U}P_{V}} m_{ij} s_{ij}} \big] \\
& - \mathbb{E}_{(\mathbf{u}_i, \mathbf{v}_i)\sim P_{U}P_{V} } \big[ \frac{N s_{ii}}{s_{ii} + (N-1) \sum_{j = 1}^{N}\mathbb{E}_{(\mathbf{u}_i, \mathbf{v}_j) \sim P_{U}P_{V}} m_{ij} s_{ij}}  \big] + 1 \Big\}  \\
= & \max_{ \{m_{ij}\} } \mathbb{E}_{(\mathbf{u}_i, \mathbf{v}_i)\sim P_{UV} } \big[\log \frac{s_{ii}}{s_{ii} + (N-1) \sum_{j = 1}^{N}\mathbb{E}_{(\mathbf{u}_i, \mathbf{v}_j) \sim P_{U}P_{V}} m_{ij} s_{ij} }\big]  + C \\
= & \max_{ \{m_{ij}\} } \mathbb{E}_{i \in \mathcal{V} } \big[\log \frac{s_{ii}}{s_{ii} + (N-1) \sum_{j \in \mathcal{V}} m_{ij} s_{ij} } \big] + C \\
= & - \min_{ M } \mathbb{E}_{i \in \mathcal{V} } \big[ \mathcal{L}_{\text{NML}} ^{(i)} \big] + C = I_{\text{NML}} (U;V), 
\end{aligned}
\end{equation*}
\end{tiny}
where $C = \log N$. Therefore, $I_{\text{NML}} (U; V)$ is a lower bound of $I ( U; V )$. 

\textbf{(2) Tighter than InfoNCE. }
Then we will analyze the relationship between our loss and InfoNCE loss.  
\begin{equation*}
\begin{aligned}
I_{\text{NML}} = & \max_{M } \mathbb{E}_{i \in \mathcal{V} } \big[\log \frac{s_{ii}}{s_{ii} + (N-1) \sum_{j \in \mathcal{V}} m_{ij} s_{ij} } \big] + C \\
\ge & \mathbb{E}_{i \in \mathcal{V} } \big[ \log \frac{s_{ii}}{s_{ii} + \sum_{j \in \mathcal{V}, j\ne i} s_{ij} } \big]  + C \\
= & - \mathbb{E}_{i \in \mathcal{V} } \big[ \mathcal{L}_{\text{InfoNCE}} \big]  + C = I_{\text{NCE}}, 
\end{aligned}
\end{equation*}
where the inequality holds because InfoNCE simply sets $m_{ij} = 0$ if $i=j$ otherwise $m_{ij} = \frac{1}{N - 1}$ while $I_{\text{NML}} $ seeks the optimum $m_{ij} \in [0,1]$. Therefore, the optimization of negative metric network $M$ gives a tighter estimation of $I(U; V)$ than InfoNCE. In conclusion, $I (U; V) \ge I_{\text{NML}} (U; V)  \ge I_{\text{NCE}}(U; V)$ holds.
\end{proof}

\subsection{Proof of Theorem \ref{th:relation}}\label{proof:th2}
\begin{proof}
Let $T$ be the variable of labels in the graph $\mathcal{G}$, and $e_1, e_2$ be the random perturbation variables introducing by the graph augmentation strategies, which are mutually independent. As the randomness of views $\mathcal{G}_U$ and $\mathcal{G}_V$ are caused by the random perturbations $e_1, e_2$, $\mathbb{E}_{\mathcal{G}_U, \mathcal{G}_V} [ I( U ; V ) ] = \mathbb{E}_{e_1, e_2} [ I( U ; V ) ]$, and thus the optimal $E^* = \argmax_{E} \mathbb{E}_{\mathcal{G}_U, \mathcal{G}_V} [ I( U ; V ) ] = \argmax_{E} \mathbb{E}_{e_1,e_2} [ I( U ; V ) ]$.

Since in GCL, the augmentations $\mathcal{G}_U$ and $\mathcal{G}_V$ are first generated by random perturbations $e_1, e_2$ and the labels $T$ in $\mathcal{G}$, and then fed to encoder $E$ to get the embeddings $U$ and $V$, Markov Chains can be constructed to describe this process, which are: (1) $\{ T, e_1 \} \to {\mathcal{G}_u} \to U$ and (2) $\{ T, e_2 \} \to {\mathcal{G}_v} \to V$. According to the Data Processing Inequality \cite{alajaji2018introduction}, we can get an upper bound of $I(U; V)$ as 
\begin{equation*}
\begin{aligned}
I (U; V) \le I( \mathcal{G}_U ; \mathcal{G}_V ). 
\end{aligned}
\end{equation*}
From this upper bound, we can see that if one optimizes $E$ by maximizing $\mathbb{E}_{e_1, e_2} [ I( U ; V ) ]$, the optimal $E^*$ would be lossless, i.e, $\mathbb{E}_{e_1, e_2} [ I( U^* ; V^* ) ] = \mathbb{E}_{e_1, e_2} [ I( \mathcal{G}_U ; \mathcal{G}_V ) ]$ where $U^* = E^* (\mathcal{G}_U)$ and $V^* = E^* (\mathcal{G}_V)$. 

By the chain rule of mutual information, i.e., $I(A;B,C) = I(A; B) + I(A; C \vert B)$, we have 
\begin{small}
\begin{equation*}
\begin{aligned}
& \mathbb{E}_{e_1, e_2} \big[ I( \mathcal{G}_U ; \mathcal{G}_V ) \big] = \mathbb{E}_{e_1, e_2} \big[ I( \mathcal{G}_U ; T , e_2 ) \big] \\
=& \mathbb{E}_{e_1, e_2} \big[ I( \mathcal{G}_U ; T) + I( \mathcal{G}_U ; e_2 \vert T ) \big] = \mathbb{E}_{e_1, e_2} \big[ I( T , e_1 ; T ) + I( T , e_1 ; e_2 \vert T ) \big] \\
=& \mathbb{E}_{e_1, e_2} \big[ I( T ; T ) + I( T ; e_1 \vert T ) + I( T ; e_2 \vert T ) + I( e_1 ; e_2 \vert T ) \big] , 
\end{aligned}
\end{equation*}
\end{small}
where $I( T ; e_1 \vert T ) =0$ and $I( T ; e_2 \vert T ) = 0$ hold obviously. And since $e_1, e_2$ are independent, $I( e_1 ; e_2 \vert T ) = 0$. We can further get that 
\begin{equation*}
\begin{aligned}
\mathbb{E}_{e_1, e_2} \big[ I( \mathcal{G}_U ; \mathcal{G}_V ) \big] = \mathbb{E}_{e_1, e_2} \big[ I( T ; T ) \big] = I( T ; T ), 
\end{aligned}
\end{equation*}
from which we can see that, if the encoder $E$ is optimized by $\max_{E} \mathbb{E}_{e_1, e_2} [ I( U ; V ) ]$, the optimal $E^*$ satisfies $\mathbb{E}_{e_1, e_2} [ I( U^* ; V^* ) ] = I(T; T)$, which indicates that $E^*$ can simulate to the oracle label function $Y$ in Definition \ref{Def:false-negative} and thus $\forall i \in \mathcal{V} , j \in \mathcal{S}_{i}^{+}$, $z_i = z_j$ holds. 

Therefore, by the properties of distance metric $d(\cdot, \cdot)$, we further have 
\begin{equation*}
\begin{aligned}
\mathbb{E}_{i \in \mathcal{V}, j \in \mathcal{S}_{i}^{+}} [d(z_i, z_j)] = 0 \le \mathbb{E}_{i \in \mathcal{V}, j \in \mathcal{S}_{i}^{-}} [d(z_i, z_j)], 
\end{aligned}
\end{equation*}
i.e., $E^*$ can accurately rank the false and true negatives since the label of any anchor node and its false negatives are the same while the label of an anchor node and its true negatives are different. 
\end{proof}

\section{Additional Experiments} \label{addExp}

\subsection{False Negative Identification}
Fig. \ref{fig:fn_tn_weights} illustrates how the learned weights of negatives ($m_{ij}$ in Equation (\ref{eq:m})) change with the number of iterations during training of NML-GCL on six datasets, where red curve represents the sum of the weights of true negatives, and blue curve represents the sum of the weights of false negatives. We can see that as the number of iterations increases, the weights of false negatives gradually decrease, while the weights of true negative examples gradually increase. This indicates that, in the negative metric space learned by NML guided by the self-supervised signal, false negatives  are gradually moving closer to the anchor nodes, while true negatives are moving further away, which enhances the false negative recognition capability of NML-GCL. 
% The weight curves on the rest datasets are shown in Fig. \ref{fig:fn_tn_weights_appendix} in Appendix, which gives similar results.

We further investigate the distribution of similarity between negatives and anchor nodes before and after applying NML-GCL. The first row of Fig. \ref{fig:negative_dist} shows the similarity distribution computed using the original node features on six datasets, while the second row of Fig. \ref{fig:negative_dist} shows the similarity distribution computed using the node embeddings generated by NML-GCL. We can see that before applying NML-GCL, the similarity distributions of true negatives and false negatives with the anchor nodes are overlapping and difficult to distinguish. However, after applying NML-GCL, the two distributions become clearly separated. Specifically, since NML-GCL brings false negatives closer to the anchor nodes while pushing true negatives farther away, the distribution of false negatives shifts toward higher similarity (rightward) while the distribution of true negatives shifts toward lower similarity (leftward). 
% In Appendix, Fig. \ref{fig:sim_appendix} shows similar results on the rest datasets.

\subsection{Verification of Theory Analysis}
\begin{figure}[t]
    \centering
    \includegraphics[width=0.95\linewidth]{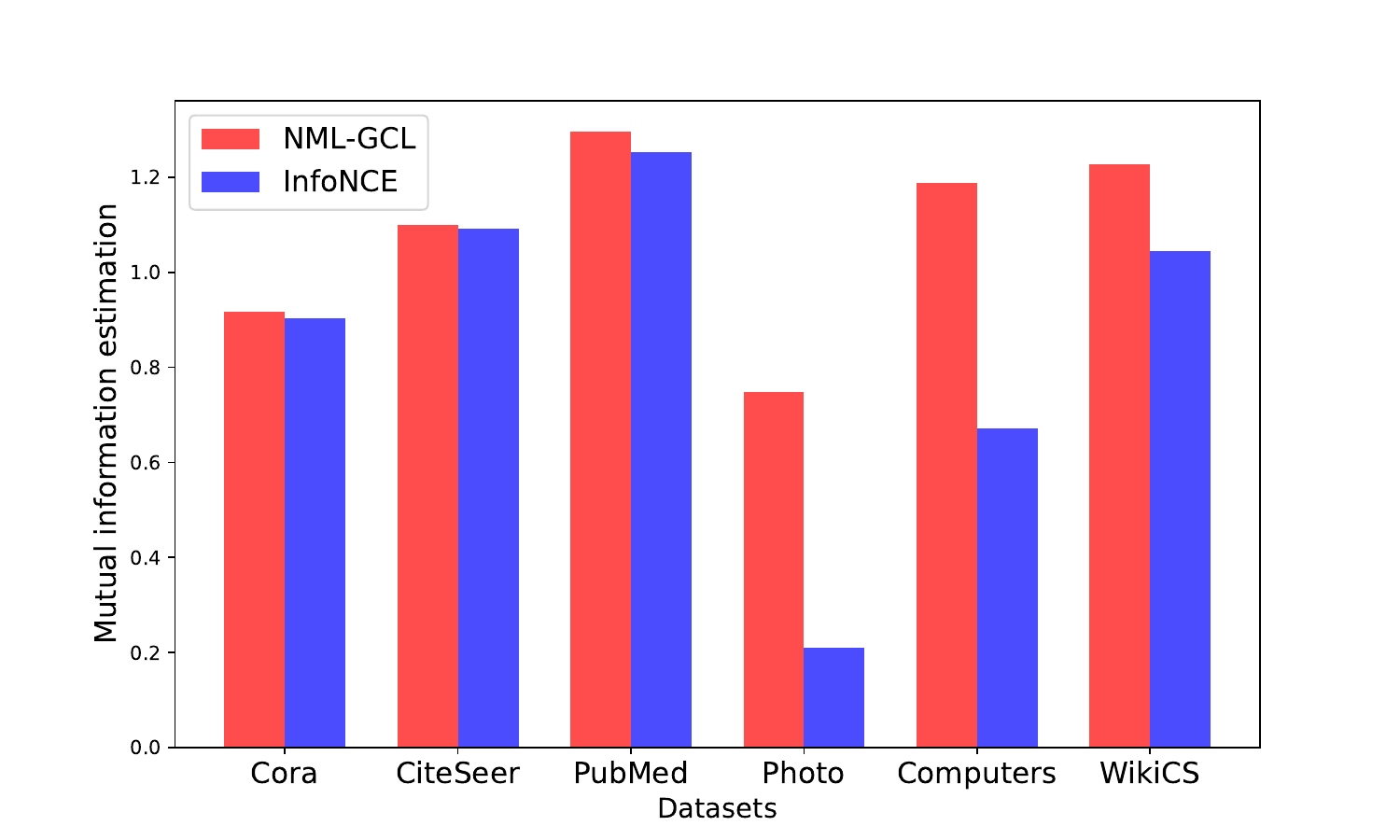}
    \caption{Mutual information estimation.}
    \label{fig:mi_comparison}
\end{figure}

In this section, we experimentally verify the theoretical conclusions of Theorem \ref{th:tighter} and Theorem \ref{th:relation}. Fig. \ref{fig:mi_comparison} shows the average mutual information estimated at each iteration during contrastive learning. We can see that on each dataset, the MI $I_{\text{NML}} (U; V)$ estimated by our contrastive loss $\mathcal{L}_{\text{NML}}^{(i)}$ defined in Equation (\ref{eq:cons_loss}) is bigger than $I_{\text{NCE}}(U; V)$ estimated by InfoNCE loss defined in Equation (\ref{eq:infonce}), which is consistent with the conclusion of Theorem \ref{th:tighter}. We also present in Table \ref{tab:distance} the ratio of the median distance of false negatives and true negatives to anchor nodes. From Table \ref{tab:distance} we see that on each dataset, NML-GCL results in the smallest ratio, indicating that, compared to the baseline methods, NML-GCL can pull false negatives closer to an anchor node while pushing true negatives farther away from it, which is consistent with the conclusion of Theorem \ref{th:relation}. Interestingly, from Table \ref{tab:distance}, we can also observe that each of the three categories of baseline methods has two runners-up. This indicates that there is no clear advantage among them in separating false negatives from true negatives, which stands in stark contrast to NML-GCL.

\section{Time Complexity Analysis} \label{time}
 we provide a brief analysis of the time complexity of the training algorithm (Algorithm \ref{algorithm}), where we take the number of nodes $N$ in the graph as the input size. From Algorithm \ref{algorithm}, we can see that the basic operation of the innermost loop is the computation of Equation (\ref{eq:objective}) (line 8). The time complexity of computing Equation (\ref{eq:objective}) in the worst case is $O(N^2)$, since it requires calculating the loss mean across all nodes. Therefore, the time cost of the innermost loop (lines 7-9) is $O(T_M N^2)$, and the update in line 12, which involves the computation of Equation (\ref{eq:objective}), still requires $O(N^2)$. Therefore, for one iteration of the outer loop, the time complexity is $O(T_M N^2 + N^2)$, and with $T_E$ times of iterations in the outer loop, the total time complexity is $O(T_E (T_M N^2 + N^2)) = O(N^2)$, as $T_E$ and $T_M$ are constants.

\section{Hyper-parameter Setting} \label{hyper-appendix}
Table \ref{tab:datasets} shows in detail the statistical characteristics of the used datasets and the hyper-parameter settings of NML-GCL on the corresponding datasets. Our graph augmentation is uniformly achieved by removing 40\% of the edges and masking 10\% of the attributes.
For baseline methods, the grid-search is used for seeking optimal hyper-parameters to achieve a fair comparison, as shown in Table \ref{tab:hyper-setting}. 

\begin{table*}[t]
  \centering
    \begin{tabular}{l|cccccc}
    \toprule
    \textbf{Method} & \textbf{Cora} & \textbf{CiteSeer} & \textbf{PubMed} & \textbf{Photo} & \textbf{Computers} & \textbf{Wiki-CS} \\
    \midrule
    BGRL & 92.25 & 96.49 & 91.20 & \underline{62.23} & 71.98 & 95.30 \\
    GRACE & 89.50 & 95.04 & 91.40 & 60.30 & 63.26 & \underline{74.20} \\
    MVGRL & 82.05 & 90.48 & 85.71 & 100.28 & 100.26 & 99.37 \\
    \midrule
    LOCAL-GCL & 92.30 & 96.70 & 92.22 & 77.30 & 71.95 & 80.58 \\
    PHASES & \underline{72.10} & 91.21 & \underline{87.56} & 70.99 & 72.12 & 78.75 \\
    HomoGCL & 83.24 & 89.33 & 91.10 & 100.15 & 100.51 & 99.21 \\
    \midrule
    GRACE+ & 87.73 & 90.69 & 91.23 & 68.15 & 64.82 & 77.03 \\
    ProGCL & 82.74 & \underline{87.80} & 93.49 & 62.30 & \underline{60.79} & 89.40 \\
    GRAPE & 91.98 & 89.35 & OOM & 99.92 & 100.26 & 87.28 \\
    \midrule
    NML-GCL & {\textbf{71.23}} & {\textbf{85.72}} & {\textbf{84.13}} & {\textbf{52.81}} & {\textbf{60.25}} & {\textbf{71.02}} \\
    \bottomrule
    \end{tabular}%
    \caption{The ratio (\%) of the median distance of false negatives and true negatives to anchor nodes. The best result on each dataset is highlighted in boldface and the runner-up results are underlined.}
  \label{tab:distance}%
\end{table*}%

\begin{table*}[t]
  \centering
    \begin{tabular}{l|cccc|cccccc}
    \toprule
    \textbf{Dataset} & \textbf{\#Nodes} & \textbf{\#Edges} & \textbf{\#Attributes} & \textbf{\#Classes} & learning rate & weight decay & $\tau$ & $T_E$ & $T_M$ & $\alpha$ \\
    \midrule
    Cora  & 2,708  & 10,556 & 1,433  & 7 & 5e-4 & 1e-3 & 0.8 & 200 & 2 & 0.1 \\
    CiteSeer & 3,327  & 9,228  & 3,703  & 6 & 5e-4 & 5e-3 & 0.7 & 50 & 3 & 0.1 \\
    PubMed & 19,717 & 88,651 & 500 & 3 & 5e-4 & 0 & 0.5 & 150 & 3 & 0.05\\
    Photo & 7,650  & 119,081 & 745 & 8 & 1e-4 & 0 & 0.5 & 50 & 5 & 0.1\\
    Computers & 13,752 & 245,861 & 767 & 10 & 5e-4 & 0 & 0.4 & 50 & 8 & 0.1\\
    Wiki-CS & 11,701 & 216,213 & 300 & 10 & 5e-4 & 0 & 0.5 & 50 & 6 & 0.2\\
    \bottomrule
    \end{tabular}%
  \caption{Statistics of the used datasets and corresponding hyper-parameter settings of NML-GCL.}
  \label{tab:datasets}%
\end{table*}%

\begin{table*}[t]
  \centering
    \begin{tabular}{l|l|l}
    \toprule
    \textbf{Method} & \textbf{Non-shared Hyper-parameters} & \textbf{Shared Hyper-parameters} \\
    \midrule
    BGRL & \makecell[c]{$\eta \in \{0.00001, 0.0001, 0.0005\}$,\\  $\text{training-epochs} \in [1, 2000]$} & \multirow{12}{*}{\makecell[l]{$\text{feature-perturbation-ratio} \in \{0.1, 0.2, 0.3, 0.4, 0.5\}$, \\$\text{edge-perturbation-ratio} \in \{0.1, 0.2, 0.3, 0.4, 0.5\}$, \\ $\text{learning-rate} \in \{0.0001, 0.0005,  0.001, 0.005, 0.01, 0.1\}$, \\ $\text{weight-decay} \in \{0.0001, 0.0005,  0.001, 0.005, 0.01, 0.1\}$, \\$\tau \in [0.1, 1]$}} \\\cmidrule{1-2}
    GRACE &  \makecell[c]{$\text{training-epochs} \in \{200, 1000, 1500, 2000, 2500\}$} \\\cmidrule{1-2}
    MVGRL &  \makecell[c]{$\text{batch-size} \in \{2, 4, 8\}$, \\ $\text{training-epochs} \in [1, 2000]$} \\\cmidrule{1-2}
    LOCAL-GCL &  \makecell[c]{$\text{steps} \in \{20, 40, 50\}$, $d \in \{1024, 2048\}$} \\\cmidrule{1-2}
    PHASES & \makecell[c]{$K_m \in \{50, 500, 1000\}$, \\ $\text{training-epochs} \in \{200, 1000, 1500, 2000, 2500\}$} \\\cmidrule{1-2}
    HomoGCL &  \makecell[c]{$\text{n-clusters} \in \{10, 30\}$, \\ $\text{training-epochs} \in \{50, 80, 100, 150\}$} \\\cmidrule{1-2}
    GRACE+ &  \makecell[c]{$f_{weight} \in \{0.5, 0.8\}$, \\ $\text{training-epochs} \in \{500, 1000, 1500, 2000\}$} \\\cmidrule{1-2}
    ProGCL &  \makecell[c]{$\text{training-epochs} \in \{200, 1000, 1500, 2000, 2500\}$} \\\cmidrule{1-2}
    NML-GCL &  \makecell[c]{$T_E \in \{50, 100, 150, 200\}, T_M \in [1, 10], \alpha \in [0, 1]$}  \\
    \bottomrule
    \end{tabular}%
    \caption{Hyper-parameter search spaces of baseline methods and NML-GCL.}
  \label{tab:hyper-setting}%
\end{table*}%

\end{document}